\documentclass[accepted]{uai2025} 
                        
\usepackage[american]{babel}

\usepackage{natbib} 
\usepackage{mathtools} 
\usepackage{booktabs} 
\usepackage{tikz} 

\usepackage{bm}
\usepackage{amsmath}
\usepackage{amssymb}
\usepackage{mathtools}
\usepackage{amsthm}
\usepackage{algorithm, algpseudocode}
\usepackage{float}
\usepackage{subcaption}
\usepackage{soul}
\usepackage{caption}
\usepackage{multirow}
\usepackage{xcolor}

\DeclareMathOperator*{\argmax}{argmax}

\newtheorem{lemma}{Lemma}

\newtheorem{definition}{Definition}

\title{A Trajectory-Based Bayesian Approach to Multi-Objective Hyperparameter Optimization with Epoch-Aware Trade-Offs}

\author[1]{Wenyu~Wang}
\author[2,3]{Zheyi~Fan}
\author[1]{Szu~Hui~Ng}
\affil[1]{%
    Department of Industrial Systems Engineering and Management, National University of Singapore, Singapore
}
\affil[2]{%
    Academy of Mathematics and System Sciences, Chinese Academy of Sciences, China
}
\affil[3]{%
    School of Mathematical Sciences, University of Chinese Academy of Sciences, China
}

\begin{document}
\maketitle

\begin{abstract}
    Training machine learning models inherently involves a resource-intensive and noisy iterative learning procedure that allows epoch-wise monitoring of the model performance. However, the insights gained from the iterative learning procedure typically remain underutilized in multi-objective hyperparameter optimization scenarios. Despite the limited research in this area, existing methods commonly identify the trade-offs only at the end of model training, overlooking the fact that trade-offs can emerge at earlier epochs in cases such as overfitting. To bridge this gap, we propose an enhanced multi-objective hyperparameter optimization problem that treats the number of training epochs as a decision variable, rather than merely an auxiliary parameter, to account for trade-offs at an earlier training stage. To solve this problem and accommodate its iterative learning, we then present a trajectory-based multi-objective Bayesian optimization algorithm characterized by two features: 1) a novel acquisition function that captures the improvement along the predictive trajectory of model performances over epochs for any hyperparameter setting and 2) a multi-objective early stopping mechanism that determines when to terminate the training to maximize epoch efficiency. Experiments on synthetic simulations and hyperparameter tuning benchmarks demonstrate that our algorithm can effectively identify the desirable trade-offs while improving tuning efficiency.
\end{abstract}

\section{Introduction}\label{sec-introduction}

With the expanding complexity of machine learning (ML) models, there is a significant surge in the demand for Hyperparameter Optimization (HPO). This surge is not only in pursuit of model prediction accuracy but also for ensuring the computational efficiency and robustness of models in real-world scenarios, which leads to the optimization task of finding the trade-off hyperparameter settings among multiple competing objectives $\bm{f} = \{f_1, \dots,f_k\}$, known as Multi-Objective Hyperparameter Optimization (MOHPO) \citep{eggensperger2021hpobench,karl2023multi,morales2023survey}. 
While MOHPO focuses solely on tuning hyperparameters $\bm{x}$, we extend this framework at its core by jointly tuning $\bm{x}$ and training epochs $t$, i.e., effectively optimizing objectives $\bm{f}(\bm{x}, t)$, to uncover superior trade-offs that emerge during iterative training (see Section \ref{sec-EMOHPO} for more details).

\begin{figure*}
    \centering
    \includegraphics[width=\textwidth]{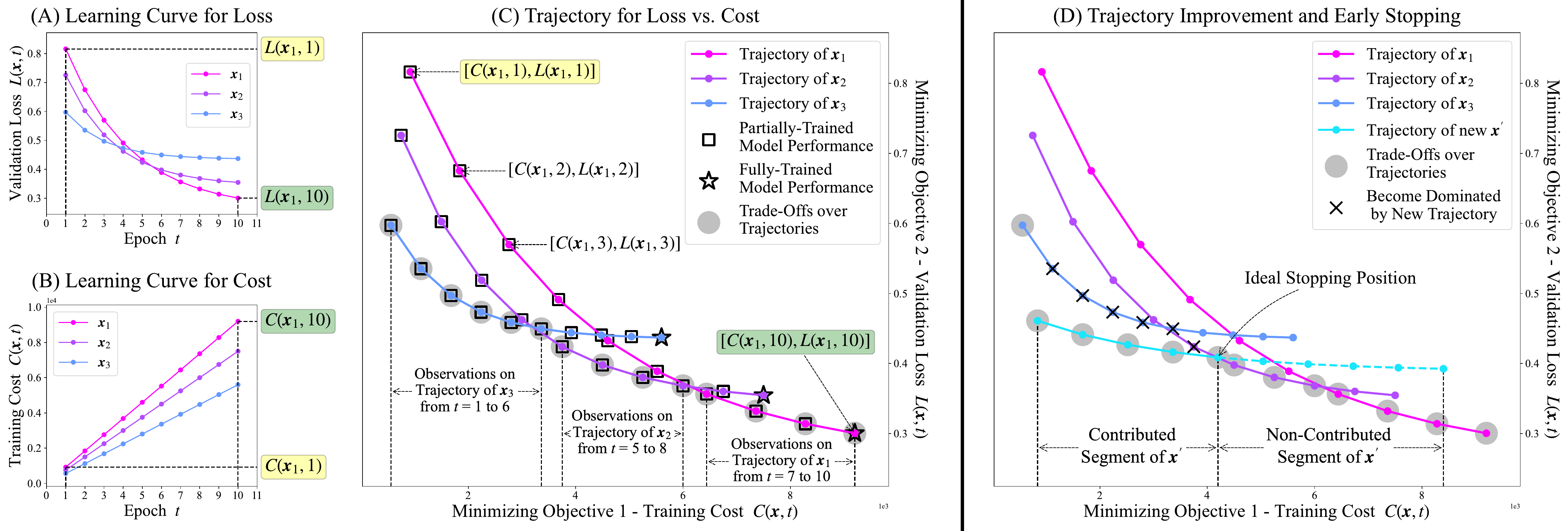}
    \caption{(A) and (B): Learning curves of three hyperparameter settings $\bm{x}_1$, $\bm{x}_2$, and $\bm{x}_3 \in \mathbb{R}^d$; (C): Trajectories of $\bm{x}_1$, $\bm{x}_2$, and $\bm{x}_3$ and trade-offs over their model performances; (D) Trajectory-based improvement and early stopping when a new $\bm{x}' \in \mathbb{R}^d$ is sampled. $L(\bm{x}, t)$ (or $C(\bm{x}, t)$) denotes the validation loss (or cost) of training with $\bm{x} \in \mathbb{R}^d$ for $t$ epochs. The maximum number of epochs is 10.}
    \label{fig-example}
\end{figure*}

Addressing HPO has long been challenging as it involves resource-intensive model training that prevents optimizers from exhaustively exploring the hyperparameter space. In this context, Bayesian Optimization (BO) has become increasingly popular \citep{srinivas2009gaussian,bergstra2011algorithms,levesque2016bayesian,foldager2023role}. This approach builds a probabilistic surrogate model, e.g., Gaussian Process (GP) \citep{williams2006gaussian}, for the objective and samples a new solution by maximizing an acquisition function formulated by the prediction and uncertainty of the surrogate model. 
Nevertheless, traditional BO methods require observing the model performance that is fully trained after a maximum number of epochs, which could potentially lead to a waste of computational resources if early indications suggest sub-optimal performance. In general, the training behind many ML models is an iterative learning procedure where a gradient-based optimizer updates the model epoch by epoch. This procedure allows users to delineate a learning curve for any hyperparameter setting by epoch-wisely monitoring the intermediate model performance (see Figure \ref{fig-example}(A) and (B)), benefiting from which prior research has introduced a set of epoch-efficient single-objective BO methods \citep{swersky2014freeze,dai2019bayesian,nguyen2020bayesian,belakaria2023bayesian} to avoid computational waste. 

The concept of leveraging iterative learning to achieve more granular control over training epochs can also be extended to MOHPO. Existing methods in multi-fidelity MOHPO \citep{belakaria2020multi,schmucker2021multi} offer a relevant perspective by treating training epoch as an auxiliary parameter and aim to use the information obtained from lower fidelities (i.e., fewer training epochs) to facilitate the search for the trade-offs at the highest fidelity (i.e., the maximum training epoch). The key assumption underlying this stream of research is that higher fidelity levels generally lead to more desirable model performance. However, this assumption does not always hold in the context of ML training. For example, in the case of overfitting, a model trained for fewer epochs can outperform a fully-trained model, making it the desired outcome. This further leads to a critical question: \textit{does a trade-off among multiple objectives emerge when the number of training epochs is fewer than the maximum allowed?}

Figure \ref{fig-example}(C) depicts the objective space of an HPO with two objectives (training cost and validation loss), where each $d$-dimensional hyperparameter setting $\bm{x}$ is trained for up to $10$ epochs, with its fully-trained model performance denoted by a star symbol. To better capture these dynamics, we extend the concept of the learning curve by introducing the notion of a ``trajectory'' to describe the evolution of model performance across epochs during the iterative training procedure. For instance, the trajectory of $\bm{x}_1 \in \mathbb{R}^d$ (denoted by the pink curve) consists of its model performances observed from epoch $t = 1$ to $10$. The emergence of trajectories highlights the limitation of multi-fidelity MOHPO in that their optimality is limited to fully-trained model performances only, ignoring a large amount of partially-trained model performances ($t < 10$) that can also contribute to trade-offs. 

To address the aforementioned limitation, in this study, we propose to jointly tune the hyperparameter setting and training epoch in the context of iterative learning, and formulate a novel Enhanced MOHPO (i.e., EMOHPO) whose optimization target is to uncover the trade-offs across all partially- and fully-trained model performances, or equivalently, the \textit{trade-offs over trajectories} (depicted by the shaded points in Figure \ref{fig-example}(C)). In fact, one important application of EMOHPO is to avoid the overfitting issue. The trade-offs of EMOHPO provide valuable insights for determining how many epochs should be allocated to achieve better generalization. Additionally, in scenarios where ML models are periodically retrained on similar datasets and need rapid deployment, such as recurrent data analyst jobs in the cloud \citep{casimiro2020lynceus,mendes2020trimtuner} and data drift detection in self-adaptive systems \citep{mahadevan2024cost,casimiro2024self}, the decision maker can benefit from the trade-offs of EMOHPO to gain a better understanding of the optimal hyperparameter setting and training epoch to strike the desired balance between objectives while avoiding repetitive and costly tuning for each retraining cycle.

Meanwhile, it is important to note that while the training epoch in EMOHPO can be interpreted as a fidelity level, it is explicitly treated as a decision variable. This distinction makes the optimization methods designed for multi-fidelity MOHPO incompatible with EMOHPO, as they seek optimal hyperparameter settings at the highest fidelity level only and do not take training epoch as part of the decision-making process. To this end, we introduce a Trajectory-based MOBO (i.e., TMOBO) algorithm, which is designed to fully leverage the trajectory information for sequential sampling and granular control over training epochs. More specifically, TMOBO samples a hyperparameter setting in each iteration, as is common in BO methods; however, we introduce a specialized acquisition function that accounts for the contribution of the entire trajectory of model performances, instead of any single one of them, associated with a hyperparameter setting. This extension is crucial as each trajectory may contribute to multiple trade-offs (illustrated in Figure \ref{fig-example}(C)). Then, during the iterative learning of the sampled hyperparameter setting, TMOBO epoch-wisely updates predictions for the unobserved segment of the trajectory and decides on termination to improve algorithm efficiency. The proposed early stopping mechanism additionally ensures that the iterative learning continues until sufficient trade-offs along the trajectory have been collected, making it significantly different from the stopping criteria used in previous studies.

\textbf{Contributions:} (1) For the first time we formulate EMOHPO to account for the evolution of model performance across epochs and hence uncover trade-offs over trajectories which are often overlooked in multi-fidelity MOHPO studies. (2) We introduce a trajectory-based MOBO method to solve EMOHPO. The proposed method samples the next hyperparameter setting using a novel acquisition function that encapsulates trajectory information and determines when to terminate the iterative learning by a conservative early stopping mechanism. (3) Through comprehensive experiments on synthetic simulations and machine learning benchmarks, we demonstrate the effectiveness and efficiency of our method in identifying trade-offs while conserving computational resources.

\section{Enhanced Multi-Objective Hyperparameter Optimization}\label{sec-EMOHPO}

Throughout the paper, we consider the sequential minimization of an Enhanced Multi-Objective Hyperparameter Optimization problem (EMOHPO) formulated as follows,
\begin{align}
    \displaystyle\min_{(\bm{x}, t) \in \mathbb{X} \times \mathbb{T}} &~~\bm{f}(\bm{x}, t) = \left[f_1(\bm{x}, t), \dots, f_k(\bm{x}, t)\right], \label{eq-EMOHPO}
\end{align}
where $\bm{f}$ comprises $k$ objective functions, each of which represents a distinct performance measure of an ML model, e.g., validation loss and training cost. Each $\bm{x}$ denotes a $d$-dimensional hyperparameter setting within a compact set $\mathbb{X} \subset \mathbb{R}^d$, and $t \in \mathbb{T} = \{1, \dots, t_{max}\}$ the number of epochs used for training. Due to the iterative learning procedure, querying any feasible pair $(\bm{x}, t)$ in EMOHPO requires optimizers to sequentially observe the noisy performances $\bm{y}(\bm{x}, t') = \left[y_1(\bm{x}, t'), \dots, y_k(\bm{x}, t')\right]$ for each epoch $t' = 1$ to $t$, where $y_i(\bm{x}, t') = f_i(\bm{x}, t') + \varepsilon_i$ and $\varepsilon_i \sim \mathcal{N}(0, \sigma_i^2)$ with variance $\sigma_i^2$ for any $i \in \{1, \dots, k\}$. (For simplicity, we adopt the common assumption that the noise terms are i.i.d. across both hyperparameter settings and epochs \citep{dai2019bayesian,klein2022learning}.) In order words, querying $(\bm{x}, t)$ results in a sequence of $t$ queries from $(\bm{x}, 1)$ to $(\bm{x}, t)$, which necessitates the design of a more sample-efficient strategy to avoid redundant queries of the same hyperparameter setting with different epoch numbers, especially in decreasing order.

In the context of iterative learning, each objective function $f_i(\bm{x}, \cdot)$ with fixed hyperparameter setting $\bm{x}$ is generally viewed as a learning curve. However, a learning curve only allows for the analysis of one performance measure at a time. To comprehensively analyze multiple performance measures, we introduce the concept of trajectory for any hyperparameter setting $\bm{x}$ as the collection of all noise-free model performances during the training with $\bm{x}$, i.e., 
\begin{align*}
    Trj(\bm{x}) := \left\{\bm{f}(\bm{x}, t)\right\}_{t = 1}^{t_{max}} = \left\{\left[f_1(\bm{x}, t), \dots f_k(\bm{x}, t)\right]\right\}_{t = 1}^{t_{max}}.
\end{align*}
A trajectory inherently encapsulates the information provided by multiple learning curves. For notational convenience, let $\bm{z} = (\bm{x}, t)$ and $\mathbb{Z} = \mathbb{X} \times \mathbb{T}$. We then adopt the standard definitions of Pareto optimality for multi-objective minimization problems. 

\begin{definition}
    A solution $\bm{f}(\bm{z})$ \textit{dominates} another solution $\bm{f}(\bm{z}')$, denoted by $\bm{f}(\bm{z}) \prec \bm{f}(\bm{z}')$, if and only if (1) $f_i(\bm{z}) \leq f_i(\bm{z}')$ for all $i \in \{1, \dots, k\}$ and (2) $f_j(\bm{z}) < f_j(\bm{z}')$ for some $j \in \{1, \dots, k\}$.
    \label{def-dominance}
\end{definition}

\begin{definition}
    The \textit{Pareto-optimal set} of $\mathbb{Z}$, denoted by $Z^*$, is composed of $\bm{z} \in \mathbb{Z}$ whose $\bm{f}(\bm{z})$ is not dominated by $\bm{f}(\bm{z}')$ of any other $\bm{z'} \in \mathbb{Z}$, i.e., $Z^* = \{\bm{z} \in \mathbb{Z} \mid \nexists \bm{z}' \in \mathbb{Z}, \bm{f}(\bm{z}') \prec \bm{f}(\bm{z})\}$. The corresponding set of solutions $F^* = \{\bm{f}(\bm{z}) \mid \bm{z} \in Z^*\}$ is referred to as \textit{Pareto-optimal front}.
    \label{def-Pareto-optimal}
\end{definition}

As per Definitions \ref{def-dominance} and \ref{def-Pareto-optimal}, minimizing the EMOHPO in (\ref{eq-EMOHPO}) is equivalent to locating the Pareto-optimal set over the entire search space, except that this space is composed of all feasible pairs of hyperparameter settings and training epochs. This aligns with the purpose of this study of finding the trade-offs over trajectories. 

Finally, it is important to highlight that, beyond incorporating iterative learning, EMOHPO fundamentally differentiates itself from multi-fidelity MOHPO by including the training epoch as a decision variable. Compared to (\ref{eq-EMOHPO}), multi-fidelity MOHPO confines its search to trade-offs among the fully-trained model performances (at the highest fidelity) only and can be expressed as,
\begin{align}
    \min_{\bm{x} \in \mathbb{X}} &~~\bm{f}(\bm{x}, t_{max}) = \left[f_1(\bm{x}, t_{max}), \dots, f_k(\bm{x}, t_{max})\right], \label{eq-MOHPO}
\end{align}
where the training epoch $t_{max}$ is a fixed constant. In contrast, EMOHPO broadens the search domain from $\mathbb{X} \times \{t_{max}\}$ in multi-fidelity MOHPO to $\mathbb{X} \times \mathbb{T}$ and thus accounts for all observations on the trajectories. As a consequence, the Pareto-optimal front of EMOHPO is always superior to or at least equivalent to that of multi-fidelity MOHPO because the former additionally captures trade-offs that may emerge during iterative learning (see Figure \ref{fig-example}(C)). This broader perspective enables more efficient decision-making in hyperparameter tuning, particularly for scenarios requiring model retraining.

\section{Related Work in Bayesian Optimization}\label{sec-review}

\textbf{Bayesian Optimization for Iterative Learning:} By appropriately characterizing the learning curve, epoch-efficient BO aims to predict the fully-trained model performance based on a partially observed learning curve to avoid ineffective epochs of training. Freeze-Thaw BO \citep{swersky2014freeze} introduces GP with an exponential decaying kernel to model the validation loss over time and allows the training to be paused and later resumed under hyperparameter settings that show promise. BOHB \citep{falkner2018bohb}, a BO extension of HyperBand \citep{li2018hyperband}, allocates training epochs through random sampling and utilizes successive halving to eliminate suboptimal hyperparameter settings. Instead of dynamically allocating computational budget among hyperparameter settings, BO-BOS \citep{dai2019bayesian} combines BO with Bayesian optimal stopping to early stop the training with a hyperparameter setting predicted to yield poor model performance. Similarly, both BOIL \citep{nguyen2020bayesian} and BAPI \citep{belakaria2023bayesian} incorporate strategies for early stopping with a particular focus on considering the learning curve of the training cost. BOIL integrates the cost into the acquisition function to prioritize cost-effective ML training procedures, whereas BAPI imposes a fixed total budget over cost. Unfortunately, there has been no epoch-efficient multi-objective BO.

\textbf{Multi-Objective Bayesian Optimization:} Many MOBO methods have been developed for multi-objective HPO by extending the vanilla BO framework. These methods generally build a surrogate model for each objective to minimize the necessity of actual resource-intensive objective evaluations, and they differ in the implementation of acquisition function. A straightforward approach is to convert a multiple-objective problem into a single-objective problem through techniques like random scalarization, which allows a direct application of standard acquisition functions. For example, ParEGO \citep{knowles2006parego} and TS-TCH \citep{paria2020flexible} respectively apply Expected Improvement (EI) \citep{jones1998efficient} and Thompson Sampling (TS) \citep{thompson1933likelihood}. However, this approach often encounters limitations in adequately exploring the Pareto-optimal front. Therefore, acquisition functions biased towards the Pareto-optimal front, such as Expected Hypervolume Improvement (EHVI) \citep{emmerich2006single}, have been designed. Despite the effectiveness and popularity of EHVI, its computational intensity presents a significant challenge in the development of BO methods \citep{hupkens2015faster}. More recently, $q$EHVI \citep{daulton2020differentiable} and $q$NEHVI \citep{daulton2021parallel} have extended EHVI for parallel multi-point selection and batch optimization through Monte Carlo (MC) integration \citep{emmerich2006single}, and they have demonstrated notable empirical performance. Alternatives to EHVI can be found in \citep{hernandez2016predictive,belakaria2019max,suzuki2020multi,yang2022parallel,daulton2022multi}.

\textbf{Multi-Fidelity Bayesian Optimization:} Multi-fidelity BO has a rich research history \citep{kandasamy2016gaussian,kandasamy2017multi,sen2018multi,wu2020practical,fan2024multi} and it facilitates the optimization of fully-trained model performance by utilizing its low-fidelity approximations, which can be obtained either by using a partial training dataset or by limiting the number of training epochs. Although multi-fidelity BO shares similarities with epoch-efficient BO, it predetermines the fidelity level before initiating the iterative learning procedure and therefore ignores the observations during the procedure. For example, FABOLAS \citep{klein2017fast} considers the data subset size as the fidelity level and jointly selects the hyperparameter setting and the data subset for model training. BOCA \citep{kandasamy2017multi} expands the discrete fidelity space to continuous for a more general setting. Multi-fidelity MOBO has also been studied in the literature \citep{belakaria2020multi,schmucker2021multi}; however, it is not applicable to solving EMOHPO as defined in (\ref{eq-EMOHPO}). This is because multi-fidelity MOBO treats the training epoch merely as an additional degree of freedom, rather than part of the decision variables. Consequently, its objective remains focused on solving multi-fidelity MOHPO as defined in (\ref{eq-MOHPO}).

\section{Gaussian Process for Trajectory Prediction}\label{sec-GP}

Assume a black-box function $f$ is sampled from a GP defined by a constant zero mean function and a kernel function $K(\bm{z}, \bm{z}')$. According to GP theory, the prior distribution over any finite set of $n$ inputs $Z = \{\bm{z}_i\}_{i = 1}^n$ is a multivariate Gaussian distribution,
\begin{align*}
    f(Z) \sim \mathcal{N}(\bm{0}, K(Z, Z)),
\end{align*}
where matrix $K(Z, Z) \in \mathbb{R}^{n \times n}$ with $\left[K(Z, Z)\right]_{i, j} = K(\bm{z}_i, \bm{z}_j)$. Conditioning on the corresponding observations $Y = \{y_i\}_{i = 1}^n$ at $Z$, the posterior predictive distribution at any input $\bm{z} \in \mathbb{Z}$ is also a Gaussian distribution given by
\begin{align}
    f(\bm{z}) \mid Z, Y \sim \mathcal{N}\left(\mu(\bm{z}), \Sigma(\bm{z})\right),
    \label{eq-posterior}
\end{align}
with
\begin{align*}
    \mu(\bm{z}) &= K(\bm{z}, Z)\left[K(Z, Z) + \sigma^2 I\right]^{-1} Y, \\
    \Sigma(\bm{z}) &= K(\bm{z}, \bm{z}) - K(\bm{z}, Z)\left[K(Z, Z) + \sigma^2I\right]^{-1}K(Z, \bm{z}),
\end{align*}
where $K(\bm{z}, Z) = K(Z, \bm{z})^T \in \mathbb{R}^n$ with $\left[K(\bm{z}, Z)\right]_i = K(\bm{z}, \bm{z}_i)$. Refer to \citep{williams2006gaussian} for a comprehensive review of GPs. In each iteration of the vanilla BO method, a new input $\bm{z}' \in \mathbb{Z}$ is selected by optimizing an acquisition function derived from the predictive mean $\mu$ and uncertainty $\Sigma$. Upon observing $y'$ at $\bm{z}'$, BO advances to the next iteration with the updated input and observation sets $Z = Z \cup \{\bm{z}'\}$ and $Y = Y \cup \{y'\}$.

As each input $\bm{z} = (\bm{x}, t)$, we define the kernel function $K\left((\bm{x}, t), (\bm{x}', t')\right)$ as the product of a standard kernel $K(\bm{x}, \bm{x}')$ over hyperparameter setting and a temporal kernel $K(t, t')$ over epochs, with the latter capturing relationships across different epochs for a fixed hyperparameter setting. For instance, \citet{swersky2014freeze} proposed an exponential decaying kernel to account for the validation loss that exponentially decreases over $t$. \citet{belakaria2023bayesian} used a linear kernel when modeling learning curves related to training costs. Given that learning curves for different performance measures may exhibit different characteristics, in this study we employ specific temporal kernels if their behavior is known a prior. See Appendix \ref{appx-gpexample} for an illustrative example of GP prediction. By fitting a GP model for each objective function $f_i$, $i \in \{1, \dots, k\}$, we can predict the trajectory $Trj(\bm{x})$ for any hyperparameter setting $\bm{x}$ and further improve the accuracy of its trajectory prediction by continuously monitoring the changes in model performance during iterative learning.

\section{Trajectory-Based Bayesian Optimization Approach}

Now we introduce an epoch-efficient algorithm named Trajectory-based Multi-Objective Bayesian Optimization (i.e., TMOBO) for solving the EMOHPO as defined in (\ref{eq-EMOHPO}). This algorithm is particularly designed to efficiently navigate the trade-off model performances across multiple epochs by leveraging the insights obtained from trajectories. At its core, TMOBO features a trajectory-based acquisition function to sample hyperparameter settings and a trajectory-based early stopping mechanism to determine the number of epochs to train with each hyperparameter setting. The pseudo-code of TMOBO is presented in Algorithm \ref{algo-TMOBO}.

\begin{algorithm}[!ht]
    \caption{Framework of TMOBO}\label{algo-TMOBO}
    \textbf{Input:} Initial sets of inputs $Z$ and observations $Y$, and initial Pareto-optimal front $F$ identified from $Y$.
    \begin{algorithmic}[1]
        \While{computational budget has not been exceeded}
            \State Fit $k$ GPs with $\bm{\mu}$ and $\bm{\Sigma}$ based on sets $Z$ and $Y$. 
            \State Sample a new $\bm{x}'$ by maximizing the TEHVI acquisition function. 
            \State Initialize $Z' \leftarrow \emptyset$ and $Y' \leftarrow \emptyset$.
            \For{$t' = 1$ \textbf{to} $t_{\max}$}
                \State Continue model training for the $t'$-th epoch to obtain observation $\bm{y}(\bm{x}', t')$.
                \State Let $Z' \leftarrow Z' \cup \{(\bm{x}', t')\}$ and $Y' \leftarrow Y' \cup \{\bm{y}(\bm{x}', t')\}$ and update front $F$.
                \State Fit $k$ GPs with $\bm{\mu}$ and $\bm{\Sigma}$ based on sets $Z \cup Z'$ and $Y \cup Y'$.
                \If{EarlyStopping$(\bm{x}', t', \bm{\mu}, \bm{\Sigma}, F)$ triggered} 
                    \State Break;
                \EndIf
            \EndFor
            \State Augment $Z'$ and $Y'$ into $Z$ and $Y$ respectively. 
        \EndWhile
    \end{algorithmic}
\end{algorithm}

We initiate the algorithm by generating a set of uniformly distributed hyperparameter settings $X = \{\bm{x}_i\}_{i = 1}^{n_0}$. The training datasets, including the set of query pairs $Z$ and the set of noisy multi-objective observations $Y$, are then obtained by training the ML model with each $\bm{x} \in X$ for up to $t_{max}$ epochs. Due to the unavailability of noise-free objective values, we consider the front $F$ of the noisy observations in $Y$ as an approximate representation of the Pareto-optimal front, which is continuously updated with each new observation.

Each iteration of TMOBO is centered around a two-level sampling strategy where a hyperparameter setting $\bm{x}'$ and its corresponding number of epochs $t'$ are determined successively. The iteration starts with fitting $\bm{\mu} = \left[\mu_1, \dots, \mu_k\right]$ and $\bm{\Sigma} = \left[\Sigma_1, \dots, \Sigma_k\right]$, with $\mu_i$ and $\Sigma_i$ representing the predictive mean and uncertainty for the $i$-th objective function. An unvisited setting $\bm{x}'$ is then selected by maximizing an acquisition function that measures the potential improvement made by the trajectory of a hyperparameter setting as if it were to be fully trained. Thereafter, we train the ML model with $\bm{x}'$, monitor the model performance, and predict its future trajectory epoch by epoch. Once the criterion for early stopping is met, the training for $\bm{x}'$ is terminated to conserve the computational budget. Finally, TMOBO augments the primary datasets $Z$ and $Y$ by selecting the most informative observations associated with $\bm{x}'$. 

\subsection{Trajectory-Based Acquisition Function}\label{sec-TEHVI}

As per Definition \ref{def-hv}, Hypervolume (HV) evaluates the quality of a set of solutions in the objective space without any prior knowledge of actual Pareto-optimal front. The maximization over HV yields a set of solutions that are converged to and well-distributed along the Pareto-optimal front, which makes HV one of the most popular indicators used in multi-objective optimization. As per Definition \ref{def-hvi}, Hypervolume Improvement (HVI) is built upon HV and quantifies the increase in HV as the gain brought by a solution.

\begin{definition}
    The Hypervolume of a set of solutions $F \subset \mathbb{R}^k$ is the $k$-dimensional Lebesgue measure $\lambda$ of the subspace dominated by $F$ and bounded from above by a reference point $\bm{r} \in \mathbb{R}^k$, denoted by $HV(F \mid \bm{r}) = \lambda(\cup_{\bm{y} \in F} \left[\bm{y}, \bm{r}\right])$, where $\left[\bm{y}, \bm{r}\right]$ denotes the hyper-rectangle bounded by $\bm{y}$ and $\bm{r}$.
    \label{def-hv}
\end{definition}

\begin{definition}
    The Hypervolume Improvement of a solution $\bm{y}'$ with respect to a set of solutions $F$ and a reference point $\bm{r} \in \mathbb{R}^k$ is the increase in hypervolume caused by including $\bm{y}'$ in set $F$, denoted by $HVI(\bm{y}' \mid F, \bm{r}) = HV(F \cup \{\bm{y}'\} \mid \bm{r}) - HV(F \mid \bm{r})$.
    \label{def-hvi}
\end{definition}

Given that the objective values of any out-of-sample $\bm{z}$ are unknown ahead of time, the direct computation of HVI for $\bm{z}$ is infeasible. Therefore, Expected Hypervolume Improvement (EHVI) \citep{zitzler2007hypervolume,daulton2020differentiable} has been used in the BO framework to estimate the gain of $\bm{z}$ by taking the expectation of HVI over the predictive distribution of its objective values, i.e.,
\begin{align*}
    EHVI(\bm{z} \mid F, \bm{r}) &= \mathbb{E} \left[HVI(\bm{f}(\bm{z}) \mid F, \bm{r})\right] \notag \\
    &= \int HVI(\bm{f}(\bm{z}) \mid F, \bm{r}) \mathbb{P}(\bm{f}(\bm{z}) \mid Z, Y) d\bm{f}. 
\end{align*}
Recall that in our study each $\bm{z}$ is composed of a hyperparameter setting $\bm{x}$ and a specific epoch number $t$. Obviously, EHVI determines a hyperparameter setting $\bm{x}$ purely based on its model performance after $t$ epochs without considering any valuable insights from the past observed or future potential model performance on the trajectory $Trj(\bm{x})$. Meanwhile, due to its joint sampling of hyperparameter setting and training epochs, EHVI ignores the fact that model performances at $\bm{z}_1 = (\bm{x}, t_1)$ and $\bm{z}_2 = (\bm{x}, t_2)$ can be observed in a single query in EMOHPO. In order words, this means that redundant model training with the same hyperparameter setting is allowed, which can cause inefficiencies in the optimization of EMOHPO. 

\begin{lemma}
    Let $X_{Trj}^*$ denote the set of hyperparameter settings that belong to the Pareto-optimal set of EMOHPO, i.e., $X_{Trj}^* := \{\bm{x} \in \mathbb{X} \mid \exists t \in \mathbb{T}, \nexists (\bm{x}', t') \in \mathbb{X} \times \mathbb{T}, \bm{f}(\bm{x}', t') \prec \bm{f}(\bm{x}, t)\}$. Then, the Pareto-optimal set of EMOHPO is equivalent to the Pareto-optimal set of $X_{Trj}^* \times \mathbb{T}$.
    \label{lemma-equivalent}
\end{lemma}

The above lemma (see proof in Appendix \ref{appx-lemma}) inspires the idea of solving EMOHPO by identifying the set $X_{Trj}^*$, which consists of hyperparameter settings whose trajectories are ``best'', or more precisely, contribute to Pareto optimality, rather than directly searching for $Z^*$. By focusing on $X_{Trj}^*$ and subsequently observing the trajectory of each $\bm{x}^* \in X_{Trj}^*$, we effectively reduce the search space to a lower-dimensional domain, allowing us to determine $\bm{x}$ independently of $t$. Within the BO framework, it can be achieved by iteratively finding the trajectory that makes the most significant improvement. To this end, we introduce the Trajectory-based EHVI (TEHVI) that operates over $\bm{x}$ only and wraps $t$ into the trajectory $Trj(\bm{x})$,
\begin{align*}
    TEHVI(\bm{x} \mid F, \bm{r}) &:= \mathbb{E} \left[HVI\left(Trj(\bm{x}) \mid F, \bm{r}\right)\right] \notag \\
    &= \mathbb{E} \left[HVI\left(\Big\{\bm{f}(\bm{x}, t)\Big\}_{t = 1}^{t_{max}} \mid F, \bm{r}\right)\right].
\end{align*}
By maximizing TEHVI across the hyperparameter space, we attempt to locate the hyperparameter setting that has the best trajectory regarding the current front $F$. However, it is worth noting that TEHVI is equivalent to the joint EHVI of multiple positions along a trajectory, which has no known analytical form and becomes particularly complicated when $t_{max}$ is large. Therefore, following the previous studies on the fast computation of EHVI \citep{emmerich2006single,daulton2020differentiable}, we resort to the Monte Carlo (MC) integration for approximating TEHVI, i.e.,
\begin{align*}
    TEHVI(\bm{x} \mid F, \bm{r}) \approx \frac{1}{M} \sum_{m = 1}^M HVI\left(\widehat{Trj}_m(\bm{x}) \mid F, \bm{r}\right),
\end{align*}
where $\widehat{Tr}_m(\bm{x}) := \{\widehat{\bm{f}}_m(\bm{x}, t)\}_{t = 1}^{t_{max}}$ denotes a predictive trajectory of $\bm{x}$ sampled from the joint posterior of GPs and $M$ denotes the total number of samples. To further alleviate the computational burden, we adopt the candidate search strategy that maximizes the approximated TEHVI over a fixed-size set of candidate hyperparameter settings. Each candidate is generated by adding a Gaussian perturbation to the evaluated hyperparameter setting whose trajectory has contributed to the current front the most. It has been shown that such a candidate search guarantees asymptotic convergence to the global optimum \citep{regis2007stochastic,wang2023efficient}. More importantly, it enables the simultaneous computation of TEHVI for multiple candidates in a batch to significantly improve efficiency. Please refer to Appendix \ref{appx-algorithm detail} for more details.

\subsection{Early Stopping and Augmentation}\label{sec-early stopping and data augmentation}

The Pareto-optimal front of EMOHPO is generally composed of trajectory segments of different hyperparameter settings because trajectories can intertwine within the objective space. As illustrated by Figure \ref{fig-example}(D), the trajectory of a newly sampled hyperparameter setting $\bm{x}'$ is split into contributed and non-contributed segments where the former pushes the front forward while the latter falls into the area already dominated by the front. Intuitively, during the iterative learning procedure, as we move from the contributed towards the non-contributed segment, the procedure should be immediately terminated at the end of the contributed segment, i.e., the ideal stopping position.

However, the complete trajectory $Trj(\bm{x}')$ cannot be observed until the ML model has been fully trained with the hyperparameter setting $\bm{x}'$. To this end, we estimate a conservative stopping epoch by considering both predictive mean and uncertainty associated with the positions along the trajectory,
\begin{align*}
    t^* = \sup\big\{t \in \mathbb{T} \mid \bm{\mu}(\bm{x}', t) - \beta^{\frac{1}{2}} \bm{\Sigma}(\bm{x}', t) \prec \bm{y}, \exists \bm{y} \in F\big\},
\end{align*}
where $\beta$ is a predetermined constant controlling the confidence level. Inspired by the Lower Confidence Bound (LCB) \citep{srinivas2009gaussian}, the conservative stopping epoch is the maximum number of epochs after which any future model performance is unlikely to improve the current front with a high probability. As the iterative learning procedure proceeds by epoch, the trajectory prediction is progressively updated based on the new observations on $Trj(\bm{x}')$ and so is the conservative stopping epoch. The iterative learning procedure terminates once the current training epoch $t'$ exceeds the conservative stopping epoch $t^*$. 

\begin{figure*}
    \centering
    \includegraphics[width=\textwidth]{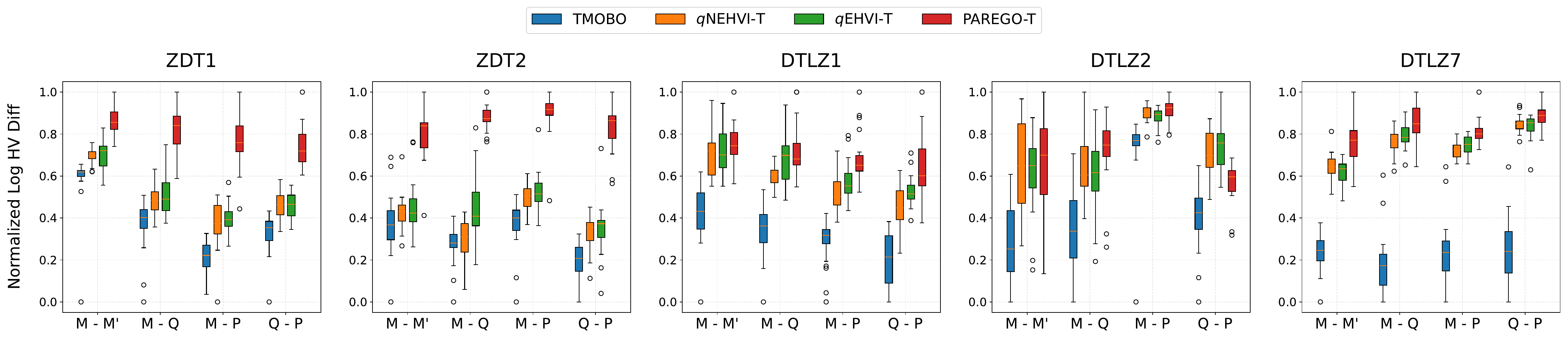}
    \caption{Box plots for 20 problems derived from ZDT1, ZDT2, DTLZ1, DTLZ2, and DTLZ7. Each algorithm runs for 20 independent trials. The logarithm of Hypervolume difference is computed at the end of each trial and is normalized in $[0, 1]$.}
    \label{fig-BoxplotSimulator}
\end{figure*}

Consequently, at the end of each iteration of TMOBO, we obtain the temporary datasets $Z' = \{(\bm{x}', t)\}_{t = 1}^{t'}$ and $Y' = \{\bm{y}(\bm{x}', t)\}_{t = 1}^{t'}$, where the actual stopping epoch $t'$ can take any value from $\{1, \dots, t_{max}\}$. However, retaining all new observations in $Z'$ and $Y'$ for training GP models in subsequent iterations is inefficient, especially when $t_{max}$ is large. Following the active data augmentation used in \citep{nguyen2020bayesian,belakaria2023bayesian}, we opt to augment only a subset of $Z'$ to the primary dataset $Z$ by sequentially selecting an input $(\bm{x}', t) \in Z'$ at which the GP predictive uncertainty is highest. As more than one GP models are used to approximate the objectives, similar to the scenario considered in \citep{belakaria2023bayesian}, the model uncertainty of $(\bm{x}', t)$ is computed as the sum of normalized variances predicted by each GP model at $(\bm{x}', t)$. Through this method, we can effectively control the increase in the size of the training set while ensuring the accuracy and training efficiency of GP models.

\section{Numerical Experiments}\label{sec-exp}

In this section, we conduct a comprehensive empirical analysis of the performance of TMOBO on several synthetic and real-world benchmark problems that are formulated as EMOHPO. Recall that EMOHPO is essentially a multi-objective optimization problem defined over the combined space of hyperparameter settings and training epochs. Therefore, we compare TMOBO with several state-of-the-art MOBO methods, namely ParEGO \citep{knowles2006parego}, $q$EHVI \citep{daulton2020differentiable}, and $q$NEHVI \citep{daulton2021parallel}. Notably, as discussed in Sections \ref{sec-introduction} and \ref{sec-review}, we exclude multi-fidelity optimization algorithms from the comparison because they do not consider the possibility that earlier epochs (or lower fidelity levels) could contribute to Pareto-optimal set or front. As a result, these methods are not applicable to solving EMOHPO. To ensure a fair comparison, we further enhance selected MOBO methods by collecting all the observations $\{\bm{y}(\bm{x}, 1), \dots, \bm{y}(\bm{x}, t)\}$ into their results when sampling at the pair $(\bm{x}, t)$. For clarity, in the following discussion, we use ParEGO-T to represent the enhanced version of ParEGO, and similarly for $q$EHVI-T and $q$NEHVI-T. 

Considering that all algorithms are stochastic, we perform 20 independent trial runs of each algorithm on each test problem. The logarithm of the HV difference between the front found by an algorithm and the true Pareto-optimal front is computed to measure the performance \citep{daulton2020differentiable,daulton2021parallel}. Since the true Pareto-optimal front is generally unknown in practice, we approximate it by aggregating all observed solutions across all algorithms and trials and extracting the non-dominated set to form an empirical Pareto front. Although the empirical front may not fully capture the true Pareto front, this ensures an unbiased comparison, as no algorithm is favored a priori, and each method’s performance is evaluated relative to the best-known solutions discovered collectively. Similarly, the reference point is set as the least favorable solution with each dimension corresponding to the worst observed value of an objective so that it upper bounds all observations in the objective space for HV computation. Further details on algorithm configurations and corresponding sensitivity analyses are provided in Appendices \ref{appx-algorithm config} and \ref{appx-sensitivity}, respectively. In the rest of this section, we showcase and analyze the main numerical results of this study but refer interested readers to Appendix \ref{appx-results} for detailed results and additional experiments.

\begin{figure*}
    \centering
    \includegraphics[width=\textwidth]{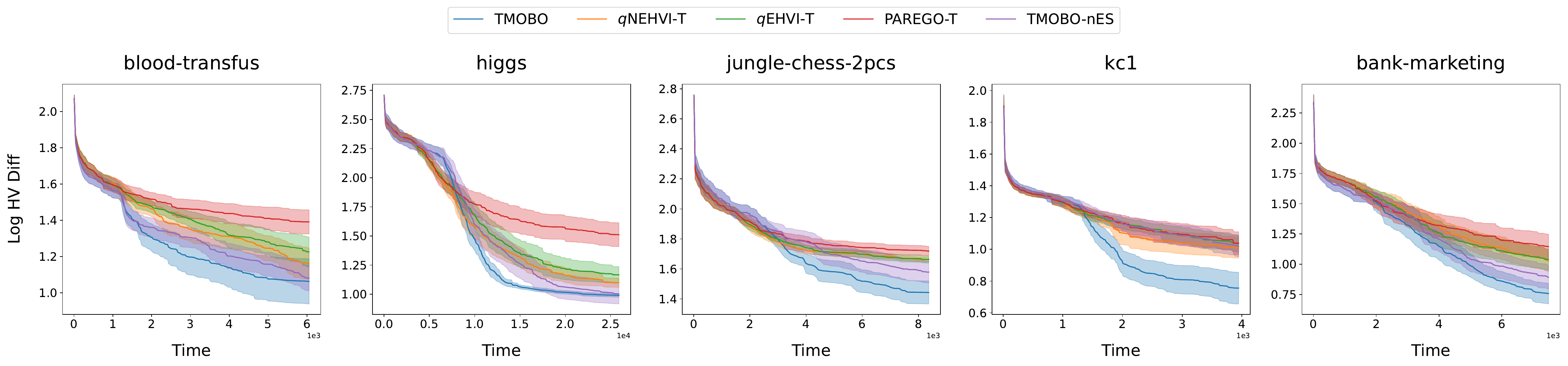}
    \caption{Average log Hypervolume difference against time for each algorithm on five different hyperparameter tuning tasks. Each algorithm runs for 20 independent trials. The shaded region indicates two standard errors of the mean.}
    \label{fig-ByTimeSurrogate}
\end{figure*}

\subsection{Synthetic Simulations}\label{sec-syntheticbenchmark}

The numerical experiments start with running TMOBO and alternative algorithms on a set of synthetic problems, through which we assess TMOBO's capability to leverage the trajectory information under different trajectory characteristics. Each synthetic problem (with objectives $[f_1(\bm{x}, t), \dots, f_k(\bm{x}, t)]$) is formulated as an epoch-dependent counterpart of a standard multi-objective benchmark problem (with objectives $[\bar{f}_1(\bm{x}), \dots, \bar{f}_k(\bm{x})]$), where $f_i(\bm{x}, t) = \bar{f}_i(\bm{x}) \cdot g_i(t)$ and curve function $g_i(t)$ emulates the iterative learning procedure. This construction approach enables us to diversify the trajectory characteristics of a synthetic problem by specifying the shape of $g_i(t)$ associated with each objective. In this study, we define the curve function $g_i(t)$ as either monotonic (M), quadratic (Q), or periodic (P) and constrain $g_i(t)$ to be positive to preserve the challenges posed by the original multi-objective benchmark problem (see Appendix \ref{appx-problem detail}).

We select five widely used multi-objective benchmark problems from ZDT \citep{zitzler2000comparison} and DTLZ \citep{deb2002scalable} test suites with $d = 5$ and add Gaussian noise with a standard deviation of $1\%$ of the range of each objective. Each subplot of Figure \ref{fig-BoxplotSimulator} depicts the box plots over four synthetic problems with different trajectory characteristics derived from the same multi-objective benchmark. For instance, ``ZDT1(M-P)'' refers to the synthetic test problem derived from ZDT1 with the first objective multiplied by the monotonic curve function and the second objective by the periodic curve function. It can be observed that with a maximum budget of $150$ iterations (or the number of times the iterative learning procedure is executed), TMOBO consistently achieves the lowest HV difference among all synthetic problems, and the solutions obtained by TMOBO generally dominate a large proportion of those obtained by alternative algorithms that do not exploit trajectory information. After being enhanced by trajectory observations, the performance of two HV-based methods, $q$NEHVI-T and $q$EHVI-T, is comparable to TMOBO on ZDT2(M-M) and ZDT2(M-Q) while ParEGO-T has the worst performance. These findings indicate that the trajectory information is beneficial to the optimization of EMOHPO and that TMOBO is able to maintain its advantages when processing trajectories with even complicated characteristics.

\subsection{Hyperparameter Tuning Benchmarks}\label{sec-mlbenchmark}

Five different Multi-Layer Perceptron (MLP) hyperparameter tuning tasks obtained from LCBench \citep{zimmer2021auto} are first utilized to examine the performance of TMOBO, where each task aims to optimize five hyperparameters (i.e., learning rate, momentum, weight decay, max dropout rate, and max number of units) by minimizing validation loss and training cost simultaneously on a specific training dataset. Following previous work \citep{falkner2018bohb,martinez2018funneled,daxberger2019mixed,perrone2018scalable}, we utilize the surrogate version of LCBench implemented in YAHPO Gym \citep{pfisterer2022yahpo} and HPOBench \citep{eggensperger2021hpobench}, where the performance metrics (i.e., objectives) are approximated by a high-quality surrogate. This approach avoids biased implementation errors, enables extensive testing, and ensures reproducibility across any environments (see Appendix \ref{appx-problem detail}). On each dataset, we observe epoch-wise model performance for any feasible hyperparameter settings up to $50$ epochs.

Figure \ref{fig-ByTimeSurrogate} compares algorithm performance over 20 independent replications, measured against cumulative model training time (excluding algorithm overhead) since model training is generally more computationally expensive and dominates hyperparameter optimization costs. This ensures a clearer understanding of how optimization efforts scale with the model complexity, thereby facilitating a more insightful comparison across different models and optimization methods. Besides the enhanced MOBO methods, we also implement a variant of TMOBO, named TMOBO-nES, which does not have an early stopping mechanism and trains each sampled setting for the maximum number of epochs. We note that TMOBO and TMOBO-nES spend more time in initialization as they train the initial settings thoroughly to capture the trajectory characteristics. Despite the initialization, given the same time budget, TMOBO achieves significantly lower HV difference than the other enhanced MOBO methods across all tasks. Meanwhile, the average performance of TMOBO surpasses TMOBO-nES, which shows the advantage of using an early stopping mechanism. While TMOBO-nES aims to learn better about trajectory characteristics through full model training, the early stopping mechanism enables TMOBO to save effort by terminating non-contributed training to explore more regions of interest.

\begin{figure}
    \centering
    \includegraphics[width=0.9\columnwidth]{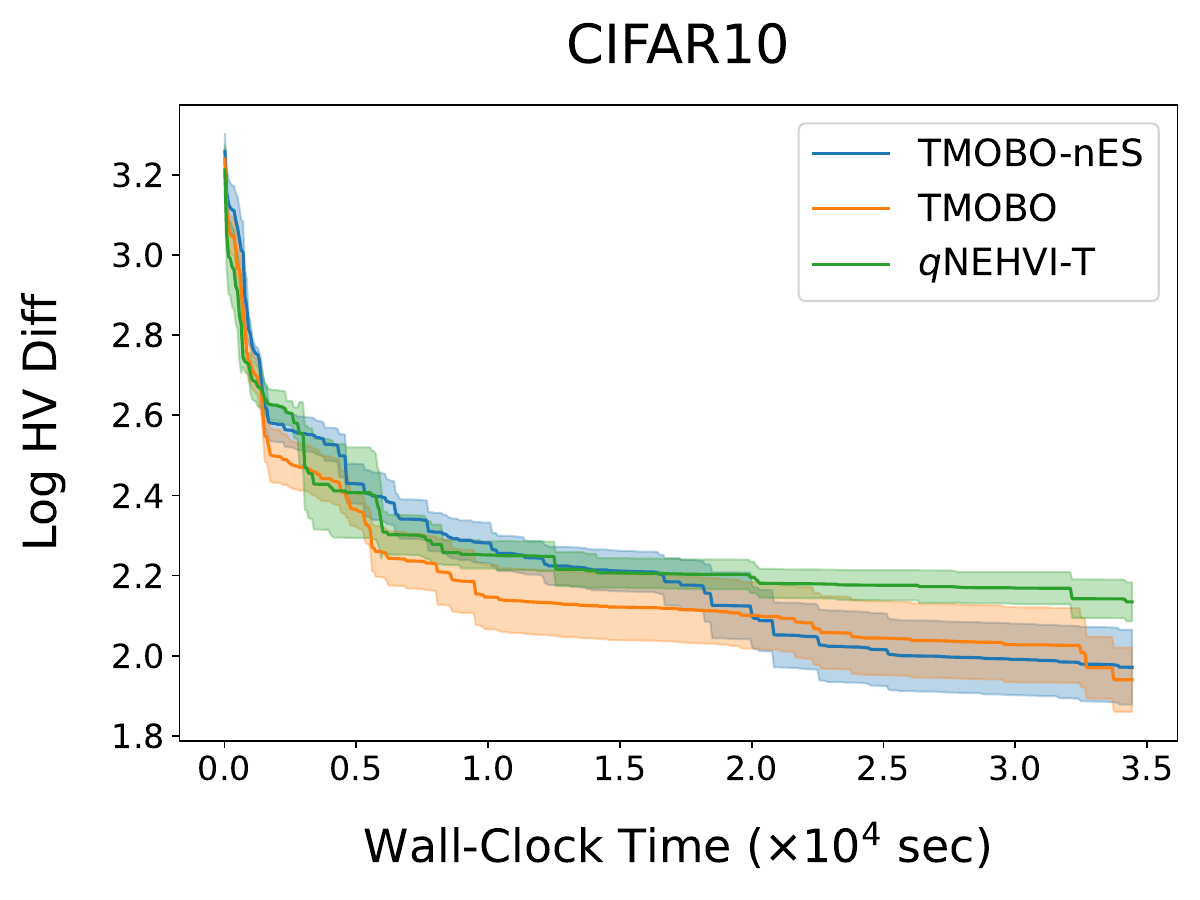}
    \caption{Average log HV difference against wall-clock time for each algorithm on the hyperparameter tuning task of MobileNetV2 on the CIFAR-10 dataset. The shaded region indicates two standard errors of the mean.}
    \label{fig-CNN Time}
\end{figure}

We further apply TMOBO, TMOBO-nES, and $q$NEHVI-T to a more challenging CNN-based task: tuning MobileNetV2 \citep{sandler2018mobilenet} on CIFAR-10 image dataset by optimizing learning rate, momentum, weight decay, batch size, and training epochs. However, unlike LCBench experiments, here we measure total wall-clock time, including CNN training and algorithm overhead. Figure \ref{fig-CNN Time} shows that after 35,000 seconds, both TMOBO and TMOBO-nES surpass qNEHVI-T in HV difference, with TMOBO converging faster early on due to early stopping. Though TMOBO-nES eventually matches TMOBO’s performance, the latter’s efficiency in the initial phase highlights its practical advantage for computationally expensive tasks. Given the inherent complexity of this CNN-based task, we believe these results provide strong evidence for the scalability and practical applicability of TMOBO.

\section{Conclusions}\label{sec-conclusions}

In this study, we consider multi-objective hyperparameter optimization with iterative learning procedures. Our interest centers on how trajectory information affects the distribution of trade-offs and on how to leverage this information to perform an effective and efficient search for trade-offs. To this end, we extend the conventional MOHPO problem to EMOHPO by including the training epoch as an explicit decision variable so as to reveal the trade-offs that may occur along trajectories. These frequently overlooked trade-offs play a beneficial role in decision-making for addressing overfitting issue and optimizing ML model deployment with retraining schemes. As there are no algorithms specially designed to handle EMOHPO, we then propose the TMOBO algorithm to first sample the hyperparameter setting with the largest trajectory-based contribution and then determine when to early stop the training with it based on the trajectory predictions of GP models. 

Through experiments on synthetic simulations and hyperparameter tuning benchmarks, TMOBO has demonstrated its advantage in locating better solutions by exploiting the trajectory information compared to traditional multi-objective optimization methods. Considering that the iterative processing procedure inherent in many real-world simulations or experiments, such as drug design and material engineering, shares similar characteristics with the training of ML models, it is meaningful to explore the formulation of EMOHPO problems across a variety of practical scenarios and to extend the success of TMOBO algorithm in this study. However, optimizing the TEHVI acquisition function remains challenging. Therefore, further exploitation is needed to derive an analytical form of TEHVI or to develop more efficient approximations.

\begin{acknowledgements} 
We gratefully acknowledge the anonymous reviewers for their valuable feedback and constructive suggestions. We would also like to thank Songhao Wang and Haowei Wang for their helpful discussions regarding this study. Szu Hui Ng’s work is supported in part by the Ministry of Education, Singapore (Grant: R-266-000-149-114).
\end{acknowledgements}

\bibliography{TMOBO}

\newpage

\onecolumn

\appendix
\setcounter{figure}{0}
\renewcommand\thefigure{\thesection.\arabic{figure}}
\setcounter{table}{0}
\renewcommand\thetable{\thesection.\arabic{table}}
\setcounter{algorithm}{0}
\renewcommand\thealgorithm{\thesection.\arabic{algorithm}}

\section{Proof for Lemma \ref{lemma-equivalent}}\label{appx-lemma}

Recall that Lemma \ref{lemma-equivalent} claims that the Pareto-optimal set of $X_{Trj}^* \times \mathbb{T}$ is equal to the Pareto-optimal set of EMOHPO, where $X_{Trj}^* = \{\bm{x} \mid \bm{x} \in \mathbb{X}, \exists t \in \mathbb{T}, \nexists (\bm{x}', t') \in \mathbb{X} \times \mathbb{T}, \bm{f}(\bm{x}', t') \prec \bm{f}(\bm{x}, t)\}$. We start by establishing the following lemma.
\begin{lemma}
    Let $Z \subset \mathbb{R}^d$ and $Z^*$ denote its Pareto-optimal set. If $Z_U$ is a subset of $Z$ such that $Z^* \subseteq Z_U \subseteq Z$, the Pareto-optimal set of $Z_U$, denoted by $Z_U^*$, is equal to $Z^*$, i.e., $Z_U^* = Z^*$.
    \label{lemma-paretosubset}
\end{lemma}

\textit{Proof.} As per Definition \ref{def-Pareto-optimal}, we have,
\begin{align}
    Z^* = \{\bm{z} \mid \bm{z} \in Z, \nexists \bm{z}' \in Z, \bm{f}(\bm{z}') \prec \bm{f}(\bm{z})\}
    \label{eq-Z*}
\end{align}
and
\begin{align}
    Z_U^* = \{\bm{z} \mid \bm{z} \in Z_U, \nexists \bm{z}' \in Z_U, \bm{f}(\bm{z}') \prec \bm{f}(\bm{z})\}. 
    \label{eq-ZU*}
\end{align}

(a) Let $\bm{z} \in Z^* \subseteq Z_U$. There does not exist $\bm{z}' \in Z$ such that $\bm{f}(\bm{z}') \prec \bm{f}(\bm{z})$. Since $Z_U \subseteq Z$, there does not exist $\bm{z}' \in Z_U$ such that $\bm{f}(\bm{z}') \prec \bm{f}(\bm{z})$. By (\ref{eq-ZU*}), $\bm{z} \in Z_U^*$ and hence $Z^* \subseteq Z_U^*$.

(b) Assume that $Z_U^* \not\subseteq Z^*$. Then, there exists $\bm{z}$ satisfying
\begin{enumerate}
    \item $\bm{z} \in Z_U^*$ $\Rightarrow$ $\nexists \bm{z}' \in Z_U$ such that $\bm{f}(\bm{z}') \prec \bm{f}(\bm{z})$; \\
    \item $\bm{z} \not\in Z^*$ $\Rightarrow$ $\exists \bm{z}'' \in Z$ such that $\bm{f}(\bm{z}'') \prec \bm{f}(\bm{z})$.
\end{enumerate}
Therefore, $\exists \bm{z}'' \in Z \setminus Z_U$ such that $\bm{f}(\bm{z}') \prec \bm{f}(\bm{z})$. Since $Z^* \subseteq Z_U$, $Z^* \cap (Z \setminus Z_U) = \emptyset$ and $\bm{z}'' \not\in Z^*$. Then, $\exists \bm{z}''' \in Z^* \subseteq Z_U$ such that $\bm{f}(\bm{z}''') \prec \bm{f}(\bm{z}'') \prec \bm{f}(\bm{z})$, which contradicts to the first condition $\bm{z} \in Z_U^*$. Therefore, $Z_U^* \subseteq Z^*$.

Combining these two inclusions together, we conclude that $Z_U^* = Z^*$. \qed

As each query pair $\bm{z} = \{\bm{x}, t\}$, the Pareto-optimal set $Z^*$ of EMOHPO, i.e., the trade-offs over trajectories, can be equivalently expressed as, 
\begin{align}
    Z^* = \{(\bm{x}, t) \mid (\bm{x}, t) \in \mathbb{X} \times \mathbb{T}, \nexists (\bm{x}', t') \in \mathbb{X} \times \mathbb{T}, \bm{f}(\bm{x}', t') \prec \bm{f}(\bm{x}, t)\}.
\end{align}
Then, we have, 
\begin{align}
    Z^* &\subseteq \{(\bm{x}, t'') \mid \bm{x} \in \mathbb{X}, \exists t \in \mathbb{T}, (\bm{x}, t) \in Z^*, t'' \in \mathbb{T}\} \notag \\
    &= \{\bm{x} \mid \bm{x} \in \mathbb{X}, \exists t \in \mathbb{T}, (\bm{x}, t) \in Z^*\} \times \{t'' \mid t'' \in \mathbb{T}\} = X_{Trj}^* \times \mathbb{T}.
\end{align}
Since $Z^* \subseteq X_{Trj}^* \times \mathbb{T} \subset \mathbb{X} \times \mathbb{T}$, we complete the proof by applying Lemma \ref{lemma-paretosubset}.

\section{Algorithm Details}\label{appx-algorithm detail}

\subsection{Gaussian Process Prediction in TMOBO}\label{appx-gpexample}

\begin{figure}[H]
    \centering
    \includegraphics[width=0.7\linewidth]{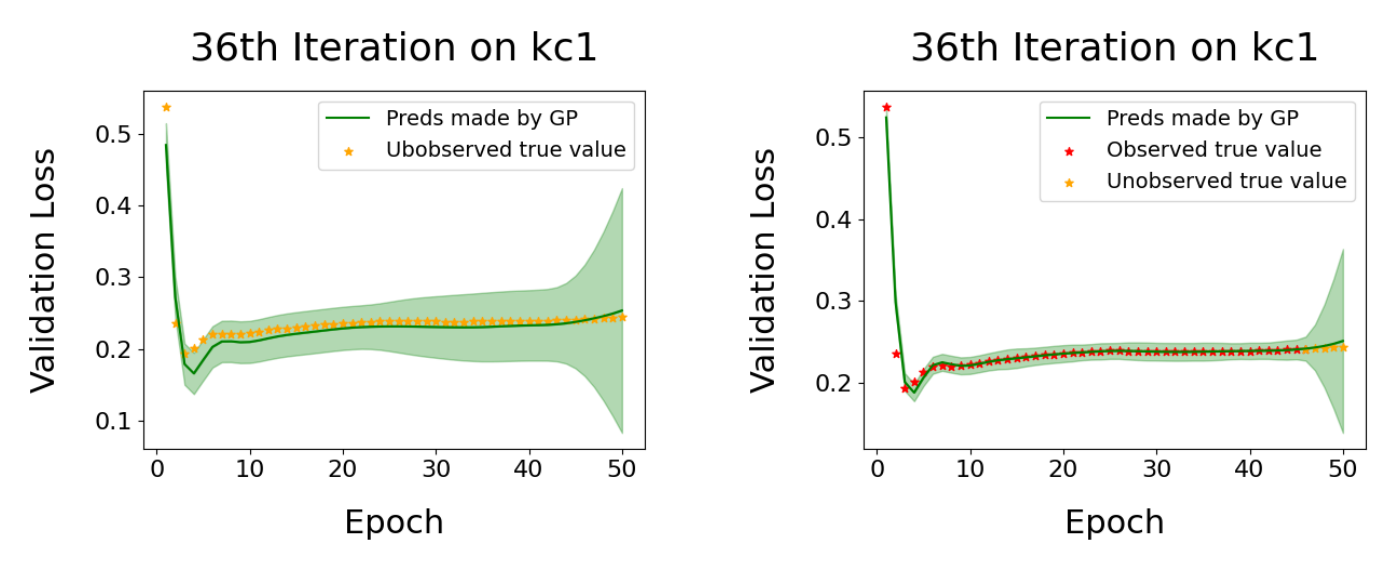}
    \caption{An illustrative example of GP prediction obtained after running TMOBO on the kc1 hyperparameter tuning task for 35 iterations. [Left] GP prediction (green curve) of validation loss for the selected hyperparameter setting over epochs 1 to 50 before any true observations (orange stars) from its learning curve are known. [Right] Updated GP predictions after some observations (red stars) on the learning curve are revealed and incorporated into the GP model.}
    \label{fig-tmobo}
\end{figure}

We begin by visualizing GP's capability of predicting learning curve at a single iteration while running TMOBO on a real hyperparameter tuning benchmark kc1. The prediction results are illustrated in Figure \ref{fig-tmobo}. Initially, even though the learning curve (depicted by orange stars) follows an uncommon shape and remains entirely unobserved, the GP model, which has been trained on data from the first 35 iterations, provides a high-quality approximation of the learning curve (depicted by the green curve) of the selected hyperparameter setting $\bm{x}$. This allows TMOBO to establish a solid foundation for trajectory prediction by leveraging multiple GP models simultaneously and computing a reliable TEHVI acquisition value before the algorithm decides to start the iterative learning process at $\bm{x}$.

As the iterative learning process progresses for the selected setting $\bm{x}$, true but noisy observations are revealed to the GP model epoch by epoch. These observations continuously enhance the predictive quality of the GP model. As shown in Figure \ref{fig-tmobo}, after incorporating the revealed observations (depicted by red stars) from the learning curve, the updated GP predictions align more closely with the true data. This refinement leads to a substantial reduction in prediction uncertainty, enabling TMOBO to make more informed decisions for early stopping.

\subsection{Candidate Search Strategy}

Acquisition function plays a critical part in the BO framework but maximizing it presents inherent challenges. As discussed in Section \ref{sec-TEHVI}, the TEHVI acquisition function extended from EHVI has no known analytical form and is multi-modal. Furthermore, approximating TEHVI by the MC method incurs significant computational overhead, making it hard to optimize efficiently. To this end, our algorithm TMOBO adopts a candidate search to determine new hyperparameter settings by strategic sampling instead of iterative optimization of the acquisition function. The candidate search first proposed by \citet{regis2007stochastic} aims to generate a set of random settings (termed as ``candidates'') around a previously visited high-quality hyperparameter setting (termed as ``center''). The new hyperparameter setting is then selected from these candidates by comparing their respective acquisition function values. This approach has demonstrated its effectiveness in global optimization \citep{regis2007stochastic} and has been extended to multi-objective cases in \citep{wang2023efficient,wang2023reference}. The specific implementation details of the candidate search in TMOBO are provided below:

\textit{Center Selection.} The effectiveness of a candidate search largely depends on the quality of the center chosen from previously visited hyperparameter settings. Intuitively, the trajectories of candidates near a high-quality center have a high chance of improving the current front $F$. However, it is inappropriate to directly use TEHVI to distinguish the best hyperparameter setting that has been visited, as trajectories of the visited hyperparameter settings have already contributed to constructing the front. Instead, similar to Def \ref{def-hvi}, we evaluate the contribution of a visited hyperparameter setting $\bm{x}$ by the difference between the HV of the front $F$ and it excluding the observed trajectory of $\bm{x}$. Therefore, the center is selected as,
\begin{align}
    \bm{x}^c = \argmax_{\bm{x} \in X} ~\big[HV\big(F \mid \bm{r}\big) - HV\big(F \setminus \big\{\bm{y}(\bm{x}, t)\big\}_{t = 0}^{t'_{max}} \mid \bm{r}\big)\big],
    \label{eq-HVC}
\end{align}
where $\bm{r}$ denotes the reference point and $\{\bm{y}(\bm{x}, t)\}_{t = 1}^{t'_{max}}$ denotes the observed trajectory when training the model with $\bm{x}$ up to $t'_{max}$ epochs. Note that $t'_{max} \leq t_{max}$ due to the early stopping mechanism.

\textit{Candidate Generation.} After selecting the center $\bm{x}^c$, we generate a candidate $\bar{\bm{x}}$ by adding Gaussian perturbation with zero mean and covariance matrix $\gamma^2(\bm{x}^c)I^d$ to the center, i.e., $\bar{\bm{x}} \sim \mathcal{N}(\bm{x}^c, \gamma^2(\bm{x}^c)I^d)$, where $\gamma(\bm{x}^c)$ denote the search radius specified for $\bm{x}^c$ with $\gamma_{max}$ being the initial value. Let $\bar{X}$ denote the set of $q$ independently generated candidates. Moreover, to dynamically balance exploration and exploitation, if the candidate search led by center $\bm{x}^c$ finally fails to yield a new setting to improve the front $F$, we halve its search radius so as to prioritize the points closer to it, and if the search fails for multiple times, we exclude $\bm{x}^c$ as a center in the subsequent iterations. Finally, we calculate the TEHVI value for each candidate in $\bar{X}$ and select the one with the highest TEHVI value to train the ML model in the current iteration.

\section{Experiment Setup}\label{appx-exp}

\subsection{Algorithm Configurations}\label{appx-algorithm config}

As discussed in Section \ref{sec-GP}, the kernel of GP models in TMOBO is implemented by the product of a Mat\'{e}rn kernel (with smoothness parameter $\nu = 2.5$) over the hyperparameter space and a temporal kernel over epochs. In real-world benchmarks, where some prior knowledge of the objective's characteristics is available, we use informed temporal kernels: an exponential decay kernel when the objective is validation loss and a linear kernel when the objective is training cost. However, in synthetic simulations, which are designed to reflect more general settings, we default to a Mat\'{e}rn temporal kernel with $\nu = 2.5$, providing a general-purpose modeling choice. The GP models are implemented using the GPyTorch library \citep{gardner2018gpytorch} and fitted via standard maximum likelihood estimation with gradient-based optimization of kernel hyperparameters.

Furthermore, within the candidate search strategy of TMOBO, the initial search radius $\gamma_{max}$ associated with any visited hyperparameter setting is set to $0.2$ as recommended by \citep{regis2007stochastic}. Moreover, to ensure a sufficiently dense neighborhood around the center, the number of candidates $q$ generated around a center is set to $100d$. Then, in the early stopping mechanism, we determine the conservative stopping epoch in a way similar to computing the lower confidence bound and maintain a fixed value of $\beta$, which controls the confidence level, at $2.0$. Finally, at the end of each iteration, TMOBO takes an active data augmentation method to minimize the prediction uncertainty of the GP model by using only a subset of trajectory observations. To maintain the GP training efficiency, we limit the size of this subset to a maximum of $10$.

We use the open-source Python implementations for ParEGO, $q$EHVI, and $q$NEHVI from the BoTorch library (accessible at \url{https://github.com/pytorch/botorch} under MIT License) and adhere to the default algorithm configurations. The enhanced versions of the alternative algorithms, namely ParEGO-T, $q$EHVI-T, and $q$NEHVI-T, are implemented by collecting all the intermediate observations $\{\bm{y}(\bm{x}, 1), \dots, \bm{y}(\bm{x}, t)\}$ to refine the current front whenever a query pair $(\bm{x}, t)$ is sampled.

In each experimental trial, we initialize each algorithm with $2(d + 1)$ samples drawn from a Sobol sequence. For the approximation of the acquisition function by Monte Carlo integration, we consistently employ $128$ MC samples across all iterations. HV-based acquisition functions in $q$EHVI, $q$NEHVI, and TMOBO are inherently sensitive to the choice of the reference point $\bm{r}$. However, determining an appropriate reference point generally requires prior knowledge of the problem, which is unrealistic for real-world applications. Thus, we adopt an adaptive strategy for all algorithms where each element of the reference point is continuously updated to the corresponding worst values encountered thus far. All experiments are run on a GeForce RTX 2080 Ti GPU with 11GB RAM.

\subsection{Problems and Benchmarks}\label{appx-problem detail}

The classical multi-objective optimization benchmarks used to formulate the synthetic test problems in numerical experiments are provided below:

\textbf{ZDT1 Benchmark} \citep{zitzler2000comparison}:
\begin{align}
    \min_{\bm{x}} ~~&[f_1(\bm{x}), f_2(\bm{x})], \notag \\
    &f_1(\bm{x}) = x_1, \\
    &f_2(\bm{x}) = u(\bm{x})\left[1 - \sqrt{x_1 / u(\bm{x})}\right], \notag
\end{align}
where 
\begin{align*}
    u(\bm{x}) = 1 + \frac{9}{d - 1} \sum_{i = 2}^d x_d,
\end{align*}
and $\bm{x} = [x_1, \dots, x_d] \in [0, 1]^d$. This benchmark has a convex Pareto-optimal front with Pareto-optimal solutions being $0 \leq x_1^* \leq 1$ and $x_i^* = 0$ for $i = 2, \dots, d$.

\textbf{ZDT2 Benchmark} \citep{zitzler2000comparison}:
\begin{align}
    \min_{\bm{x}} ~~&[f_1(\bm{x}), f_2(\bm{x})], \notag \\
    &f_1(\bm{x}) = x_1, \\
    &f_2(\bm{x}) = u(\bm{x})\left[1 - (x_1 / u(\bm{x}))^2\right], \notag
\end{align}
where
\begin{align*}
    u(\bm{x}) = 1 + \frac{9}{d - 1} \sum_{i = 2}^d x_d,
\end{align*}
and $\bm{x} = [x_1, \dots, x_d] \in [0, 1]^d$. This benchmark has a concave Pareto-optimal front with Pareto-optimal solutions being $0 \leq x_1^* \leq 1$ and $x_i^* = 0$ for $i = 2, \dots, d$.

\textbf{DTLZ1 Benchmark} \citep{deb2002scalable}:
\begin{align}
    \min_{\bm{x}} ~~&[f_1(\bm{x}), \dots, f_k(\bm{x})], \notag \\
    &f_1(\bm{x}) = \frac{1}{2} x_1 x_2 \cdots x_{k - 1} (1 + u(\bm{x}_k)), \notag \\
    &f_2(\bm{x}) = \frac{1}{2} x_1 x_2 \cdots (1 - x_{k - 1}) (1 + u(\bm{x}_k)), \\
    &~~~~~~~~~~~~\vdots \notag \\
    &f_{k - 1}(\bm{x}) = \frac{1}{2} x_1 (1 - x_2) (1 + u(\bm{x}_k)), \notag \\
    &f_k(\bm{x}) = \frac{1}{2} (1 - x_1) (1 + u(\bm{x}_k)), \notag
\end{align}
where 
\begin{align*}
    u(\bm{x}_k) = 100 \left[|\bm{x}_k| + \sum_{x_i \in \bm{x}_k}(x_i - 0.5)^2 - \cos(20\pi(x_i - 0.5))\right],
\end{align*}
$\bm{x} = [x_1, \dots, x_d] \in [0, 1]^d$ and $\bm{x}_k$ denotes the last $(d - k + 1)$ variables of $\bm{x}$. The search space of this benchmark contains multiple local Pareto-optimal fronts. The global Pareto-optimal front is a linear hyperplane with Pareto-optimal solutions being $x_i^* = 0.5$ for $x_i \in \bm{x}_k$.

\textbf{DTLZ2 Benchmark} \citep{deb2002scalable}:
\begin{align}
    \min_{\bm{x}} ~~&[f_1(\bm{x}), \dots, f_k(\bm{x})], \notag \\
    &f_1(\bm{x}) = \cos(\pi x_1 / 2) \cos(\pi x_2 / 2) \cdots \cos(\pi x_{k - 1} / 2) (1 + u(\bm{x}_k)), \notag \\
    &f_2(\bm{x}) = \cos(\pi x_1 / 2) \sin(\pi x_2 / 2) \cdots \cos(\pi x_{k - 1} / 2) (1 + u(\bm{x}_k)), \\
    &~~~~~~~~~~~~\vdots \notag \\
    &f_{k - 1}(\bm{x}) = \cos(\pi x_1 / 2) \sin(\pi x_2 / 2) (1 + u(\bm{x}_k)), \notag \\
    &f_k(\bm{x}) = \sin(\pi x_1 / 2) (1 + u(\bm{x}_k)), \notag
\end{align}
where 
\begin{align*}
    u(\bm{x}_k) = \sum_{x_i \in \bm{x}_k} (x_i - 0.5)^2,
\end{align*}
$\bm{x} = [x_1, \dots, x_d] \in [0, 1]^d$ and $\bm{x}_k$ denotes the last $(d - k + 1)$ variables of $\bm{x}$. This benchmark has a concave Pareto-optimal front with Pareto-optimal solutions being $x_i^* = 0.5$ for $x_i \in \bm{x}_k$.

\textbf{DTLZ7 Benchmark} \citep{deb2002scalable}:
\begin{align}
    \min_{\bm{x}} ~~&[f_1(\bm{x}), \dots, f_k(\bm{x})], \notag \\
    &f_1(\bm{x}) = x_1, \notag \\
    &~~~~~~~~~~~~\vdots \\
    &f_{k - 1}(\bm{x}) = x_{k - 1}, \notag \\
    &f_k(\bm{x}) = h(f_1, f_2, \dots, f_{k - 1}, u) (1 + u(\bm{x}_k)), \notag
\end{align}
where 
\begin{align*}
    &u(\bm{x}_k) = 1 + \frac{9}{|\bm{x}_k|} \sum_{x_i \in \bm{x}_k} x_i, \\
    &h(f_1, f_2, \dots, f_{k - 1}, u) = k - \sum_{i = 1}^{k - 1}\left[\frac{f_i}{1 + u} (1 + \sin(3\pi f_i))\right],
\end{align*}
$\bm{x} = [x_1, \dots, x_d] \in [0, 1]^d$ and $\bm{x}_k$ denotes the last $(d - k + 1)$ variables of $\bm{x}$. The Pareto-optimal front of this benchmark is composed of $2^{k - 1}$ disconnected regions with Pareto-optimal solutions being $x_i^* = 0$ for $x_i \in \bm{x}_k$.

Given the $i$-th objective function $f_i(\bm{x})$ of any benchmark above, we construct its epoch-dependent counterpart as $f_i(\bm{x}, t) = f_i(\bm{x}) \cdot g_i(t)$ to simulate the training procedure by the curve function $g_i(t)$. In our experiments, we utilize different curve functions to diversify the learning curve characteristics, and these functions are,
\begin{itemize}
    \item Monotonically Increasing Curve: $g^M(t \mid t_{max}) = 0.5 + 1 / \left(1 + e^{(-0.2(t - t_{max} / 2))}\right)$, \\
    \item Monotonically Decreasing Curve: $g^{M'}(t \mid t_{max}) = 0.3 + 1 / \left(1 + e^{(0.1(t - t_{max} / 3))}\right)$, \\
    \item Quadratic Curve: $g^Q(t \mid t_{max}) = 0.5 + 2\left(t / t_{max} - 2 / 3\right)^2$, \\
    \item Periodic Curve: $g^P(t \mid t_{max}) = 1 + 0.5 \sin(4\pi t / t_{max})$,
\end{itemize}
where $t \in \mathbb{T} = [1, 2, \dots, t_{max}]$ and $t_{max} = 50$. 

\textbf{LCBench} \citep{zimmer2021auto}: This benchmark is designed to give insights on multi-fidelity optimization with learning curves for Auto Deep Learning. LCBench was originally developed upon tabular data. Benefitting from the surrogate implementation by HPOBench (under Apache License 2.0) \citep{eggensperger2021hpobench} and YAHPO Gym (under Apache License 2.0) \citep{pfisterer2022yahpo}, we are allowed to observe the intermediate model performance for any feasible hyperparameter setting after each epoch. The maximum number of epochs for training MLP is set to $50$. We choose to minimize validation loss and training time, which are commonly considered in many MOHPO studies. For a demonstration, we focus on tuning the five hyperparameters (i.e., $d = 5$) of MLP including learning rate in $[1 \times 10^{-4}, 1 \times 10^{-1}]$, momentum in $[0.10, 0.99]$, weight decay in $[1 \times 10^{-5}, 1 \times 10^{-1}]$, maximum dropout rate in $[0.0, 1.0]$, and maximum number of neurons in $[64, 1024]$. LCBench utilizes diverse datasets from AutoML Benchmark \citep{gijsbers2019open} hosted on OpenML \citep{vanschoren2014openml}. For experimentation, we select ``blood-transfus'', ``higgs'', ``jungle-chess-2pcs'', ``kc1'', and ``banck-marketing'', each of which has at least two attributes and between 500 and 1,000,000 data points.

\textbf{CNN}: This benchmark provides insights into hyperparameter optimization for modern computer vision applications. The convolutional neural network model MobileNetV2 is implemented in PyTorch and optimized for both validation accuracy and training efficiency. Following standard computer vision practices, we optimize four key hyperparameters (i.e., $d = 4$), including learning rate in $[1 \times 10^{-4}, 1 \times 10^{-1}]$, momentum in $[0.0, 0.99]$, weight decay in $[1 \times 10^{-5}, 1 \times 10^{-2}]$, and batch size in $[128, 512]$. The model is trained for up to 50 epochs on CIFAR-10 dataset, which consists of 60,000 32 $\times$ 32 color images across 10 classes. This configuration represents a more complex and computationally intensive optimization scenario compared to LCBench tasks, allowing us to assess our method's scalability to challenging real-world deep learning applications.

\section{Sensitivity Analysis on Algorithm Configurations}\label{appx-sensitivity}

In this section, we conduct a series of sensitivity analyses to demonstrate how TMOBO's performance responds to its key algorithmic configurations. These configurations include the parameters for candidate search, Monte Carlo approximation, data augmentation strategy, and early stopping.

\subsection{Parameters for Candidate Search}

Regarding the candidate search, we examine both the number of candidates and the search radius. Intuitively, a larger candidate set increases the likelihood of getting promising hyperparameter settings; however, it can impose unnecessary computational costs at the same time. As shown in the third row of Figure \ref{fig-sensitivity}, using $100d$ to $200d$ candidates generally leads to a lower (better) overall distribution in the boxplots (where $d$ is the dimension of the domain of hyperparameter settings), especially when the objectives become more complex. Therefore, we choose a default setting of $100d$ candidates to balance effectiveness and efficiency. This choice is further supported by the statistical analysis in Table \ref{table-sensitivity}, which indicates that our default setting is statistically comparable to or better than other settings, except in scenario Q-P where $200d$ outperforms.

For the search radius $r$, we can observe from the last row of Figure \ref{fig-sensitivity} that a larger value of radius generally leads to a lower distribution in the boxplots; however, the marginal gains diminish gradually, and performance can even deteriorate beyond $r = 0.2$ in some cases. This behaviour is expected because the radius controls the balance between exploration and exploitation, and a larger radius biases the algorithm toward exploration by increasing the chance of selecting hyperparameter settings further from the center, which can slow convergence. Statistical results in Table \ref{table-sensitivity} indicate that $r = 0.2$ is significantly better than settings below 0.2 in half of the cases and comparable to higher settings across all cases. Consequently, we adopt it as our default.

\subsection{Parameter for Monte Carlo Approximation}

Regarding the Monte Carlo approximation, we showcase the algorithm's performance with the number of MC samples $M$ being 32, 64, 128, 256, and 512. It can be observed from the second row of Figure \ref{fig-sensitivity} that the algorithm’s performance is relatively insensitive to changes in $M$ within this range. While increasing the number of MC samples can improve the integration of predictive uncertainty, this improvement does not necessarily translate into better overall algorithm performance, but it raises the computational burden due to the expensive TEHVI acquisition function. Considering the added computational cost associated with significantly larger values of $M$, we believe that 128 samples provide a good balance, capturing sufficient information for accurate TEHVI approximation without incurring excessive computational expense. Moreover, the statistical tests in Table \ref{table-sensitivity} indicate that the algorithm performance using default setting of 128 is comparable to both lower and higher sample sizes across all tested cases.

\subsection{Parameter for Data Augmentation Strategy}

\begin{figure}
    \centering
    \includegraphics[width=0.9\textwidth]{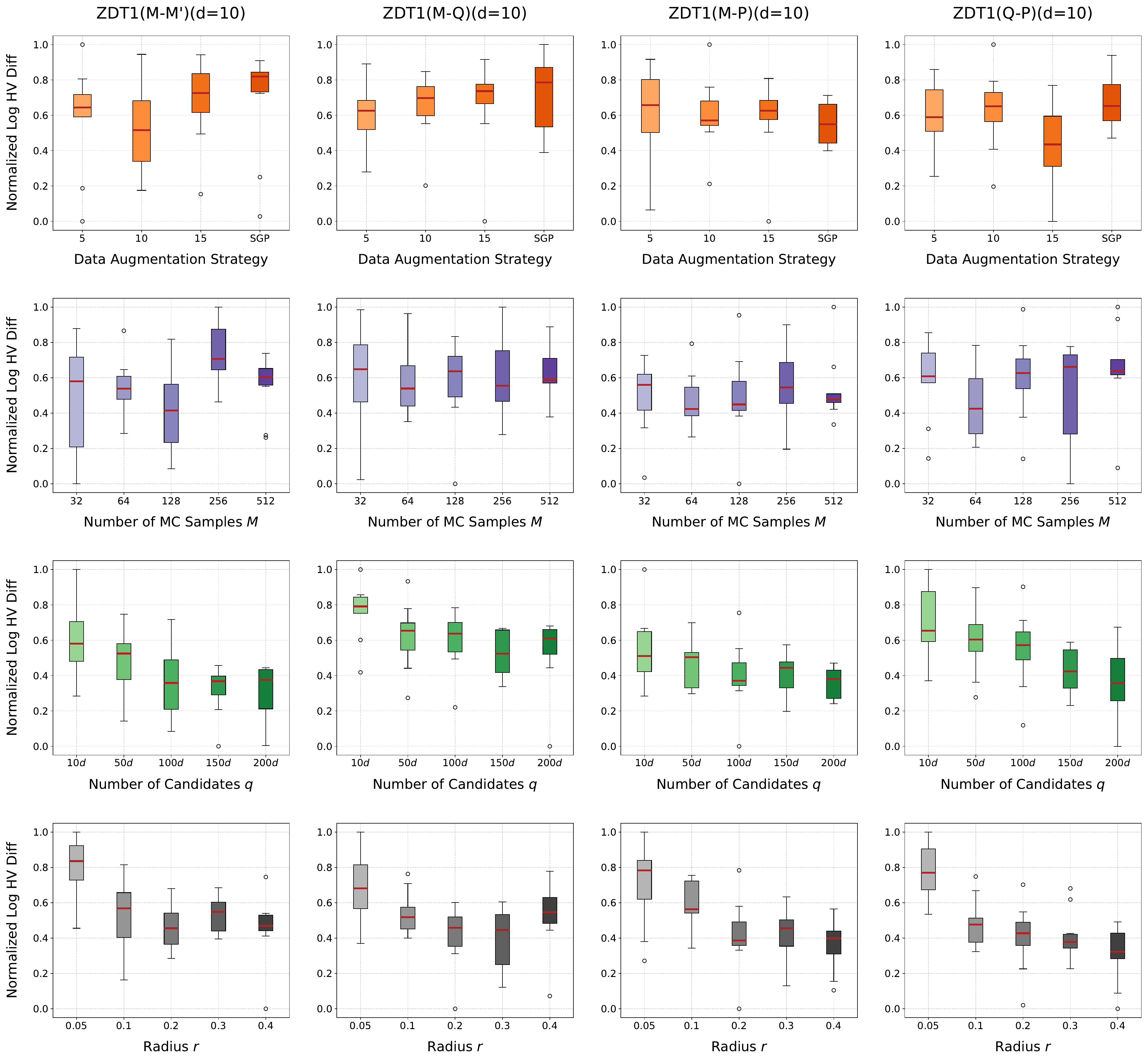}
    \caption{Sensitivity analysis on key parameters and components of TMOBO on synthetic problems generated from ZDT1 after 100 algorithm iterations. Each row varies one factor, namely (1) data augmentation strategy, using GP trained with 5, 10, or 15 observations per trajectory, or using Sparse GP (SGP); (2) number of MC samples $M \in \{32, 64, 128, 256, 512\}$; (3) number of candidates $q \in \{10d, 50d, 100d, 150d, 200d\}$, where $d$ is the problem dimension; and (4) search radius $r \in \{0.05, 0.1, 0.2, 0.3, 0.4\}$. Each box plot shows the logarithm of HV difference computed over 10 trials and is normalized to [0, 1]. (A lower value indicates better performance.)}
    \label{fig-sensitivity}
\end{figure}

We first compare data augmentation strategies by evaluating the impact of retaining 5, 10, or 15 observations per trajectory in the GP model. This analysis helps us assess whether a careful selection of augmented observations can manage GP complexity effectively while making reasonable prediction for optimization. Our findings in Figure \ref{fig-sensitivity} indicate that the algorithm’s performance is relatively insensitive to changes in this parameter across the four problems under study; in some cases, augmenting fewer data points even leads to a lower distribution in the boxplots. Considering the computational complexity of training a GP model, we maintain 10 observations per trajectory as our default setting, with the results in Table \ref{table-sensitivity} showing that this choice yields performance statistically comparable to other settings. 

In parallel, we explore scalable GP models as an alternative to manual data augmentation. To this end, we implement a sparse GP with inducing points using the GPyTorch library \citep{gardner2018gpytorch} and compare its performance against the data augmentation strategy. The Sparse GP (SGP) approach avoids the need for manual data curation and, as shown in Table \ref{table-sensitivity}, typically achieves performance comparable to the 10-observation strategy in most cases. This indicates that both a careful selection of augmented observations and the use of sparse approximations are effective means of managing GP complexity in TMOBO. Moreover, as more algorithm iterations are executed, scalable GP models can serve as a suitable alternative.

\begin{table}
    \centering
    \caption{Mean (standard deviation) of normalized log HV differences for key parameters and components of TMOBO on synthetic problems generated from ZDT1 after 100 algorithm iterations. For each parameter, performance is evaluated against the default algorithm setting (in bold) using Wilcoxon rank-sum test at a 95\% confidence level. ``$+$'' sign indicates that the parameter option performs significantly better than the baseline, ``$-$'' indicates significantly worse performance, and ``$\approx$'' denotes statistically comparable results.}
    \label{table-sensitivity}
    \resizebox{\textwidth}{!}{
    
    \begin{tabular}{cc|cccccccc}
        \hline
        & & \multicolumn{2}{c}{ZDT1(M-M')} & \multicolumn{2}{c}{ZDT1(M-Q)} & \multicolumn{2}{c}{ZDT1(M-P)} & \multicolumn{2}{c}{ZDT1(Q-P)} \\
        \cline{3-10}
         &  & Mean (std) & Sign & Mean (std) & Sign & Mean (std) & Sign & Mean (std) & Sign \\
        \hline
        \multirow{4}{*}{\parbox{3.7cm}{Data augmentation strategy}} & 5 & 0.589 (0.277) & $\approx$ & 0.603 (0.172) & $\approx$ & 0.613 (0.244) & $\approx$ & 0.587 (0.190) & $\approx$ \\
         & \textbf{10} & 0.528 (0.234) & $\approx$ & 0.659 (0.179) & $\approx$ & 0.602 (0.191) & $\approx$ & 0.628 (0.206) & $\approx$ \\
         & 15 & 0.682 (0.219) & n.a. & 0.666 (0.240) & n.a. & 0.586 (0.214) & n.a. & 0.430 (0.230) & n.a. \\
         & SGP & 0.686 (0.282) & $\approx$ & 0.722 (0.203) & $\approx$ & 0.554 (0.111) & $\approx$ & 0.684 (0.157) & $-$ \\
        \hline
        \multirow{5}{*}{\parbox{3.7cm}{Number of MC samples $M$}} & 32 & 0.493 (0.298) & $\approx$ & 0.597 (0.289) & $\approx$ & 0.496 (0.206) & $\approx$ & 0.593 (0.220) & $\approx$ \\
         & 64 & 0.552 (0.147) & $\approx$ & 0.587 (0.182) & $\approx$ & 0.470 (0.153) & $\approx$ & 0.460 (0.197) & $\approx$ \\
         & \textbf{128} & 0.417 (0.225) & n.a. & 0.583 (0.231) & n.a. & 0.486 (0.232) & n.a. & 0.603 (0.217) & n.a. \\
         & 256 & 0.734 (0.155) & $-$ & 0.616 (0.212) & $\approx$ & 0.542 (0.213) & $\approx$ & 0.498 (0.278) & $\approx$ \\
         & 512 & 0.558 (0.154) & $\approx$ & 0.631 (0.136) & $\approx$ & 0.535 (0.183) & $\approx$ & 0.651 (0.241) & $\approx$ \\
        \hline
        \multirow{5}{*}{\parbox{3.7cm}{Number of candidates $q$}} & 10$d$ & 0.600 (0.194) & $-$ & 0.761 (0.156) & $-$ & 0.545 (0.202) & $\approx$ & 0.707 (0.194) & $\approx$ \\
         & 50$d$ & 0.475 (0.196) & $\approx$ & 0.622 (0.167) & $\approx$ & 0.464 (0.142) & $\approx$ & 0.595 (0.171) & $\approx$ \\
         & \textbf{100$d$} & 0.366 (0.192) & n.a. & 0.605 (0.157) & n.a. & 0.395 (0.183) & n.a. & 0.549 (0.201) & n.a. \\
         & 150$d$ & 0.329 (0.123) & $\approx$ & 0.526 (0.129) & $\approx$ & 0.406 (0.129) & $\approx$ & 0.429 (0.122) & $\approx$ \\
         & 200$d$ & 0.314 (0.144) & $\approx$ & 0.538 (0.194) & $\approx$ & 0.356 (0.084) & $\approx$ & 0.349 (0.212) & $+$ \\
        \hline
        \multirow{5}{*}{\parbox{3.7cm}{Search radius $r$}} & 0.05 & 0.808 (0.162) & $-$ & 0.686 (0.181) & $-$ & 0.701 (0.225) & $-$ & 0.775 (0.152) & $-$ \\
         & 0.1 & 0.529 (0.205) & $\approx$ & 0.536 (0.119) & $\approx$ & 0.593 (0.132) & $-$ & 0.484 (0.132) & $\approx$ \\
         & \textbf{0.2} & 0.463 (0.120) & n.a. & 0.420 (0.167) & n.a. & 0.412 (0.190) & n.a. & 0.405 (0.175) & n.a. \\
         & 0.3 & 0.535 (0.097) & $\approx$ & 0.392 (0.162) & $\approx$ & 0.429 (0.135) & $\approx$ & 0.415 (0.129) & $\approx$ \\
         & 0.4 & 0.458 (0.177) & $\approx$ & 0.532 (0.183) & $\approx$ & 0.363 (0.142) & $\approx$ & 0.311 (0.152) & $\approx$ \\
        \hline
    \end{tabular}
    
    }
\end{table}

\subsection{Parameter for Early Stopping}

\begin{figure}[H]
  \centering
  \includegraphics[width=0.9\textwidth]{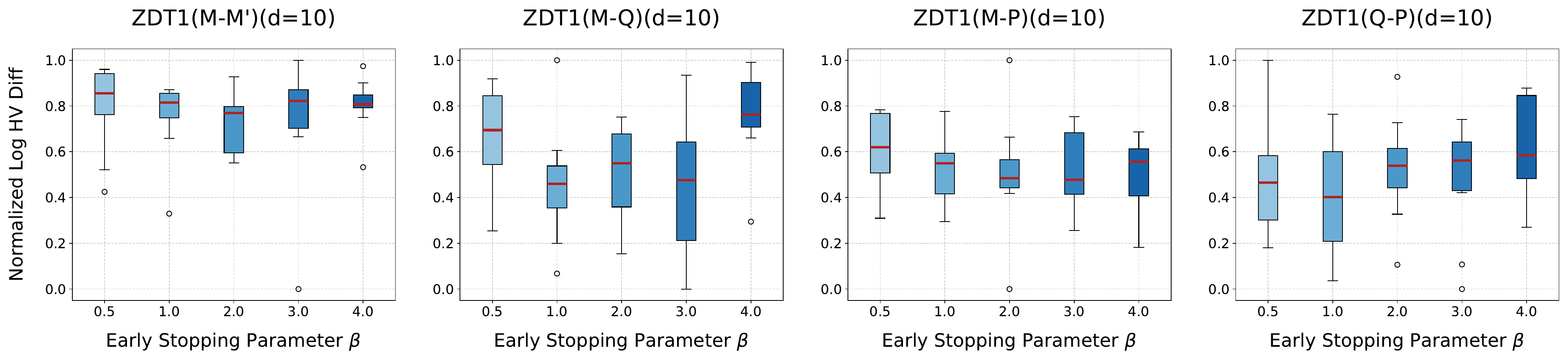}
  \caption{Sensitivity analysis on early stopping parameter $\beta \in \{0.5, 1.0, 2.0, 3.0, 4.0\}$ of TMOBO on synthetic problems generated from ZDT1 after 3000 training epochs. Each box plot shows the logarithm of HV difference computed over 10 trials and is normalized to [0, 1]. (A lower value indicates better performance.)}
  \label{fig-Boxplot-Beta}
\end{figure}

Regarding the early stopping mechanism, we present a separate sensitivity analysis in Figure \ref{fig-Boxplot-Beta} and Table \ref{table-Beta}, because the changes in $\beta$ affect the number of training epochs via early stopping, and it is more meaningful to compare the results after a fixed number of training epochs. As shown in Figure \ref{fig-Boxplot-Beta}, a moderate value of $\beta$ (between 0.1 and 0.3) generally results in a lower median in the boxplots; while $\beta = 0.05$ or $0.4$ results in a higher median, especially in the first three cases. Intuitively, a lower $\beta$ accounts for less predictive uncertainty and may cause premature termination of training, thereby encouraging exploration of more hyperparameter settings; conversely, a higher $\beta$ is likely to prolong training along a given trajectory, leading to excessive exploitation and reduced exploration. This balance between exploitation and exploration is crucial for algorithm performance. The statistical results in Table \ref{table-Beta} show that algorithm performance using the default setting $\beta = 0.2$ is statistically comparable to the other settings. Based on these insights, we recommend choosing a moderate $\beta$ value for a general TMOBO application.

\begin{table}[H]
    \centering
    \caption{Mean (standard deviation) of normalized log HV differences for early stopping parameter $\beta$ of TMOBO on synthetic problems generated from ZDT1 after 3000 training epochs. For each parameter, performance is evaluated against the default algorithm setting (in bold) using Wilcoxon rank-sum test at a 95\% confidence level. ``$+$'' sign indicates that the parameter option performs significantly better than the baseline, ``$-$'' indicates significantly worse performance, and ``$\approx$'' denotes statistically comparable results.}
    \label{table-Beta}
    \resizebox{\textwidth}{!}{
    
    \begin{tabular}{cc|cccccccc}
        \hline
        & & \multicolumn{2}{c}{ZDT1(M-M')} & \multicolumn{2}{c}{ZDT1(M-Q)} & \multicolumn{2}{c}{ZDT1(M-P)} & \multicolumn{2}{c}{ZDT1(Q-P)} \\
        \cline{3-10}
         &  & Mean (std) & Sign & Mean (std) & Sign & Mean (std) & Sign & Mean (std) & Sign \\
        \hline
        \multirow{5}{*}{\parbox{3.7cm}{Early stopping ($\beta$)}} & 0.5 & 0.802 (0.178) & $\approx$ & 0.658 (0.225) & $\approx$ & 0.606 (0.158) & $\approx$ & 0.498 (0.268) & $\approx$ \\
         & 1.0 & 0.758 (0.156) & $\approx$ & 0.460 (0.238) & $\approx$ & 0.525 (0.138) & $\approx$ & 0.404 (0.229) & $\approx$ \\
         & \textbf{2.0} & 0.735 (0.131) & n.a. & 0.496 (0.208) & n.a. & 0.502 (0.234) & n.a. & 0.530 (0.211) & n.a. \\
         & 3.0 & 0.751 (0.272) & $\approx$ & 0.459 (0.282) & $\approx$ & 0.525 (0.161) & $\approx$ & 0.481 (0.235) & $\approx$ \\
         & 4.0 & 0.805 (0.115) & $\approx$ & 0.761 (0.198) & $-$ & 0.501 (0.159) & $\approx$ & 0.618 (0.210) & $\approx$ \\
        \hline
    \end{tabular}
    
    }
\end{table}

\section{Numerical Experiments and Results}\label{appx-results}

\subsection{Additional Results for Section \ref{sec-syntheticbenchmark}}\label{appx-result-synthetic}

\begin{figure}[H]
    \centering
    \includegraphics[width=0.85\textwidth]{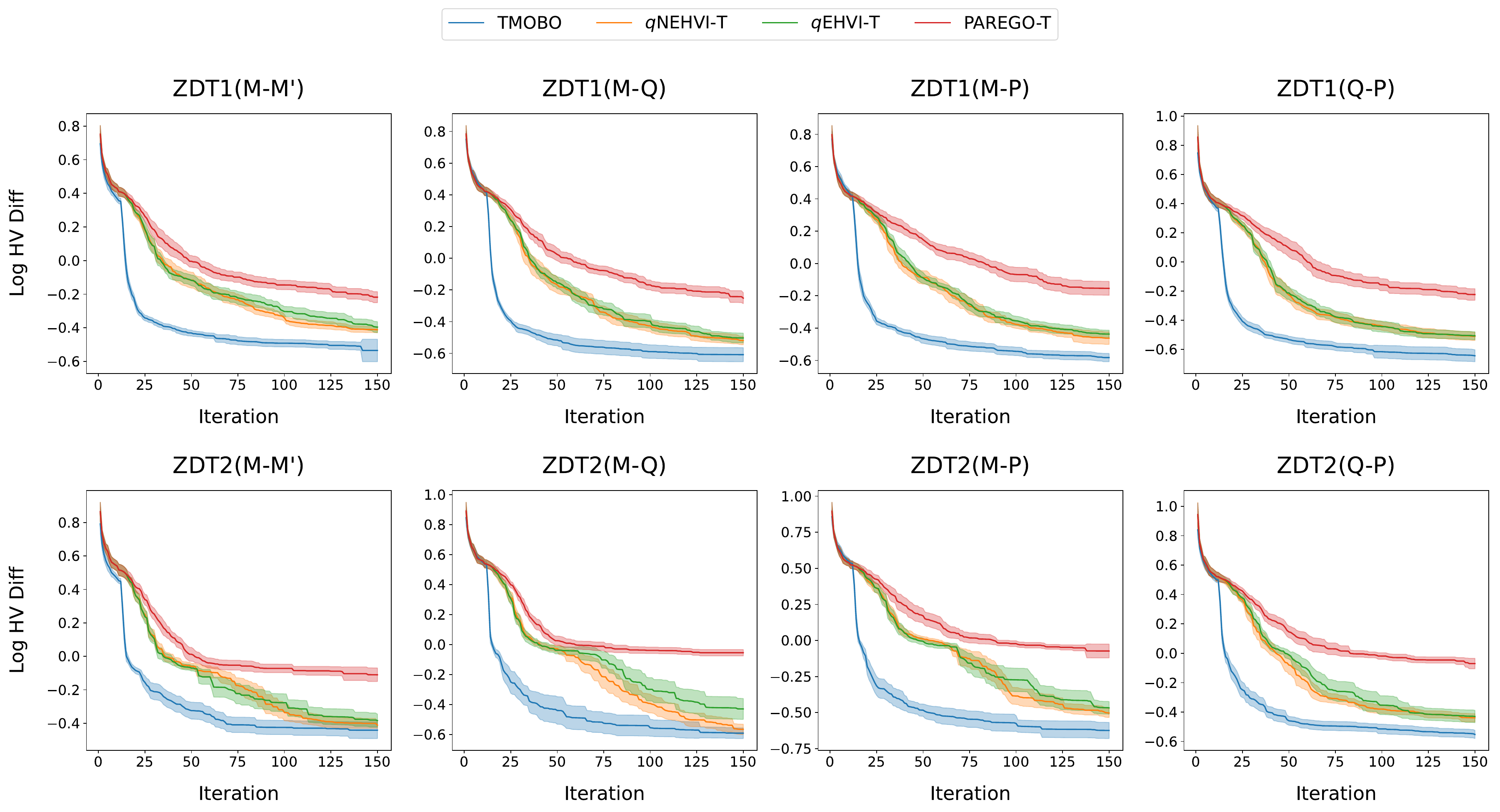}
    \caption{Average log Hypervolume difference against iterations for each algorithm on synthetic problems generated from ZDT1 and ZDT2. Each algorithm runs for 20 independent trials. The shaded region indicates two standard errors of mean.}
    \label{fig-ByIterSimulator(A)}
\end{figure}

\begin{figure}[H]
    \centering
    \includegraphics[width=0.85\textwidth]{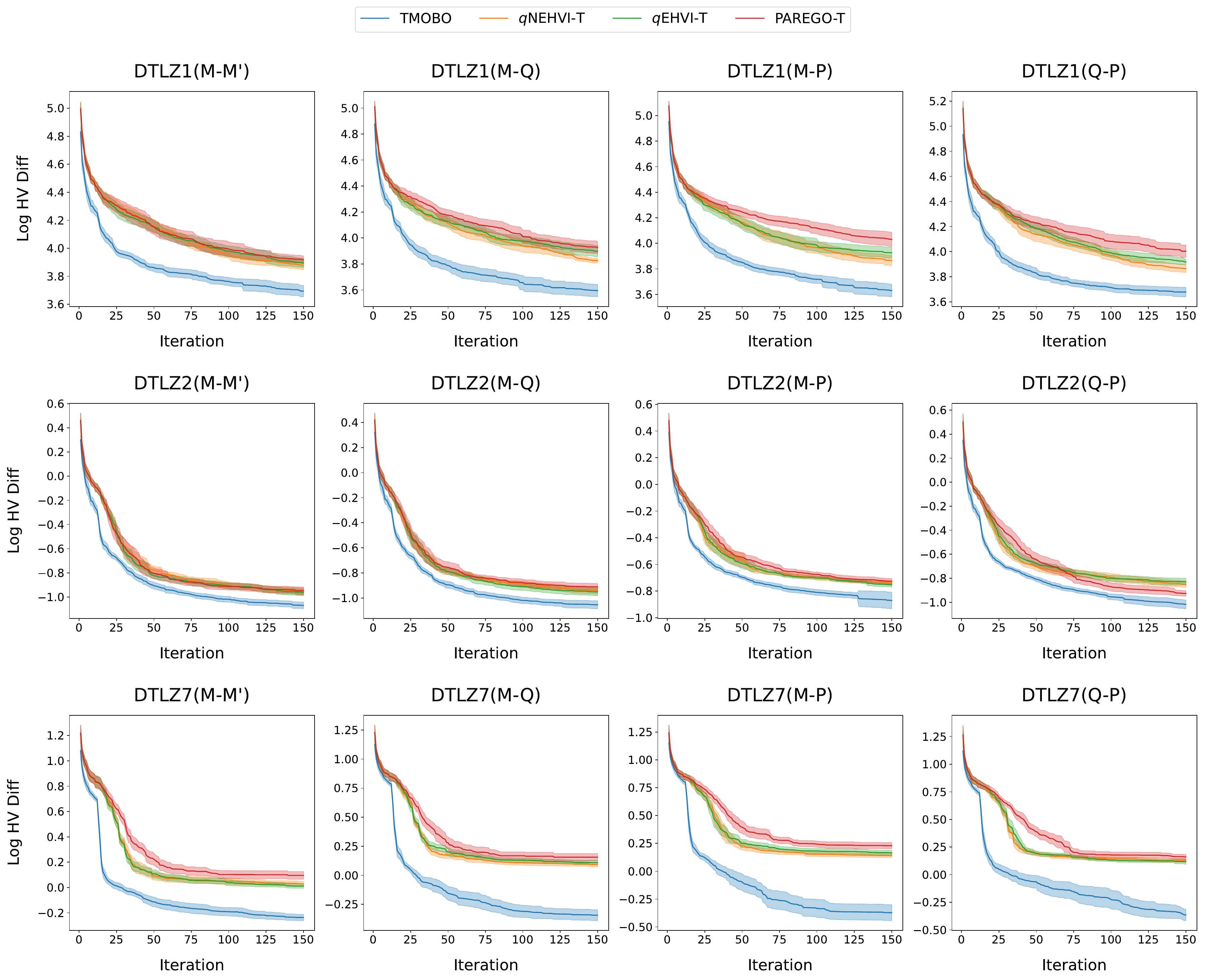}
    \caption{Average log Hypervolume difference against iterations for each algorithm on synthetic problems generated from DTLZ1, DTLZ2, and DTLZ7. Each algorithm runs for 20 trials. The shaded region indicates two standard errors of mean.}
    \label{fig-ByIterSimulator(B)}
\end{figure}

In this subsection, we evaluate TMOBO and alternative algorithms on 20 synthetic problems derived from standard multi-objective benchmarks, as discussed in Section \ref{sec-syntheticbenchmark}. Figures \ref{fig-ByIterSimulator(A)} and \ref{fig-ByIterSimulator(B)} illustrate the average log HV difference of each algorithm against the number of iterations. Each row of subplots represents problems generated from the same benchmark but with varying trajectory complexities.

On the ZDT benchmarks (Figure \ref{fig-ByIterSimulator(A)}), TMOBO achieves rapid convergence, reaching a low HV difference value within a few iterations. It maintains this advantage throughout the optimization process and outperforms the alternatives. While $q$NEHVI-T and $q$EHVI-T demonstrate competitive performance on the first two instances of ZDT2, their ability to handle increasing trajectory complexity is limited. In such cases, TMOBO significantly dominates these methods.

In contrast, the variants of the DTLZ benchmarks (Figure \ref{fig-ByIterSimulator(B)}) present greater challenges. On synthetic problems derived from DTLZ1, all algorithms struggle to converge adequately within 150 iterations, primarily due to the existence of multiple local optima. However, TMOBO achieves noticeably better HV difference values, which highlights its robustness in handling complex landscapes. DTLZ7 benchmark challenges the capability of optimizers to maintain a diverse set of solutions as it has a disconnected Pareto-optimal front. Notably, TMOBO emerges as the best choice for handling this type of challenge, as the alternatives tend to stagnate at relatively high HV difference values at an early stage.

\begin{figure}[H]
    \centering
    \includegraphics[width=\textwidth]{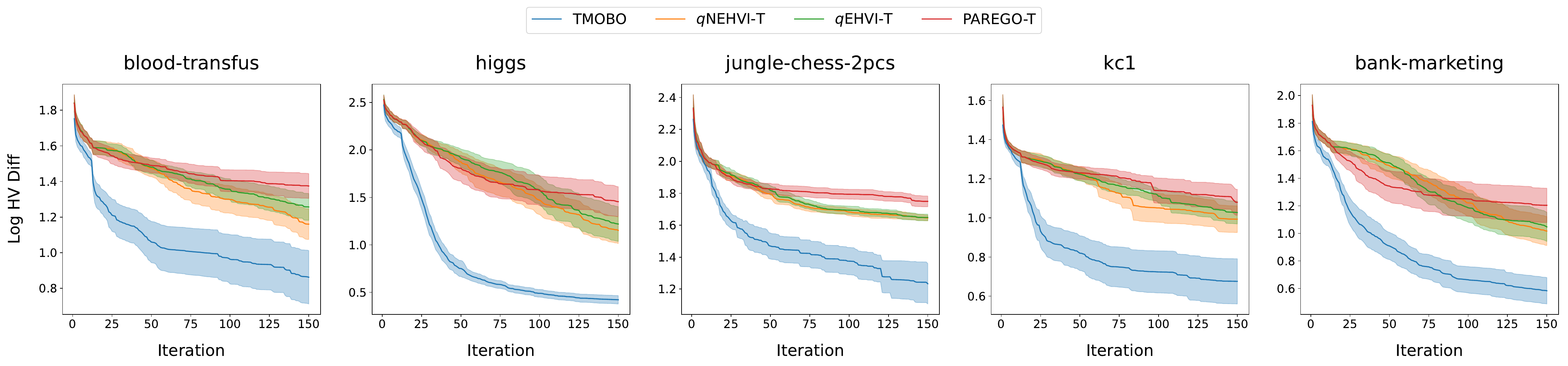}
    \caption{Average log Hypervolume difference against iterations for each algorithm on five different hyperparameter tuning tasks. Each algorithm runs for 20 independent trials. The shaded region indicates two standard errors of the mean.}
    \label{fig-ByIterSurrogate}
\end{figure}

\subsection{Additional Results for Section \ref{sec-mlbenchmark}}\label{appx-result-mlbench}

Next, we assess TMOBO and alternative algorithms on five hyperparameter tuning tasks from LCBench, as described in Section \ref{sec-mlbenchmark}. These tasks involve optimizing validation loss and training cost for an MLP model. Figure \ref{fig-ByIterSurrogate} compares the algorithms in terms of average log HV difference against the number of iterations.

TMOBO consistently outperforms the alternatives across all tasks, achieving better-converged and more diversified Pareto-optimal fronts. Its trajectory-based acquisition function enables it to extract valuable insights from partially trained models. In contrast, $q$NEHVI-T and $q$EHVI-T fail to match TMOBO's performance in maintaining convergence speed and solution quality; and ParEGO-T performs the worst. Overall, these results highlight TMOBO's effectiveness in hyperparameter tuning scenarios, particularly in balancing computational efficiency and optimization performance.

\subsection{Impact of Early Stopping}

\begin{figure}[H]
  \centering
  \includegraphics[width=0.8\textwidth]{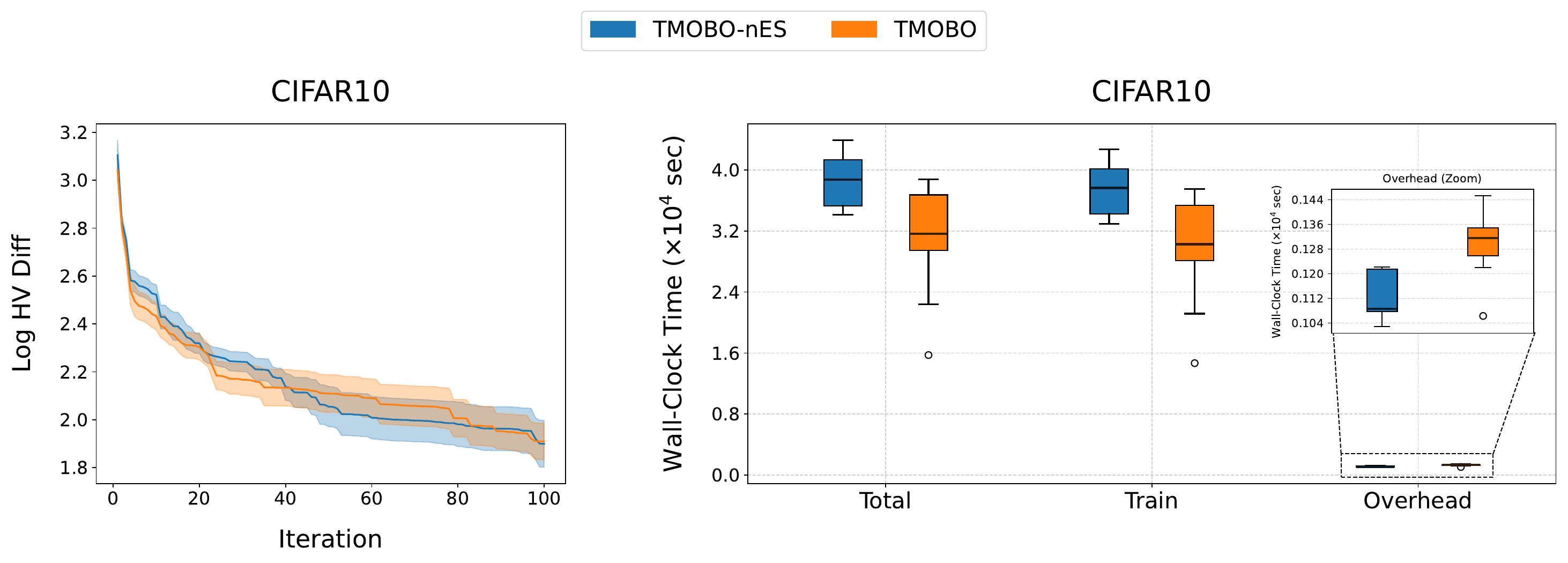}
  \caption{Comparison of the computational efficiency of TMOBO and TMOBO-nES on the hyperparameter tuning task of MobileNetV2 on the CIFAR-10 dataset. [Left] Average log HV difference against algorithm iterations for each algorithm across 10 independent trials. The shaded region indicates two standard errors of the mean. [Right] Boxplots of wall-clock time for each algorithm, where ``Total'' indicates the overall runtime, ``Train'' the overall CNN training time, and ``Overhead'' the algorithm additional processing time.}
  \label{fig-CNN}
\end{figure}

On the CNN benchmark, we further take this opportunity to provide readers with a clearer understanding of our algorithm’s early stopping mechanism as well as the computational complexity. Figure \ref{fig-CNN} first shows that TMOBO and its non-early-stopping variant, TMOBO-nES, achieve comparable performance throughout the trial, from initialization to the end of 100 algorithm iteration. Notably, TMOBO requires less overall runtime (including CNN training time and algorithm overhead) than TMOBO-nES (as reflected by the “Total” boxplots) primarily because its early stopping mechanism prevents unnecessary training of the CNN model, thereby reducing the number of epochs per trajectory and conserving the wall-clock time. This efficiency is evident in the lower “Train” boxplot of TMOBO. The trade-off is that the early stopping mechanism introduces additional computational overhead, since TMOBO needs to determine in real time when to terminate training. However, when compared with the overall time required to train the CNN model, the overhead for both TMOBO and TMOBO-nES is negligible. 

\subsection{Impact of Objective Complexity}\label{appx-result-complexity}

In different ML applications, practitioners may consider optimizing hyperparameters for varying objectives. This makes it essential to evaluate TMOBO's performance across diverse scenarios. To explore this, we modified the LCBench hyperparameter tuning tasks by replacing the training cost objective with validation cross-entropy, while retaining validation accuracy as the other objective. This modification introduces a more complex, non-linear relationship between objectives. 

\begin{figure}[H]
    \centering
    \includegraphics[width=\textwidth]{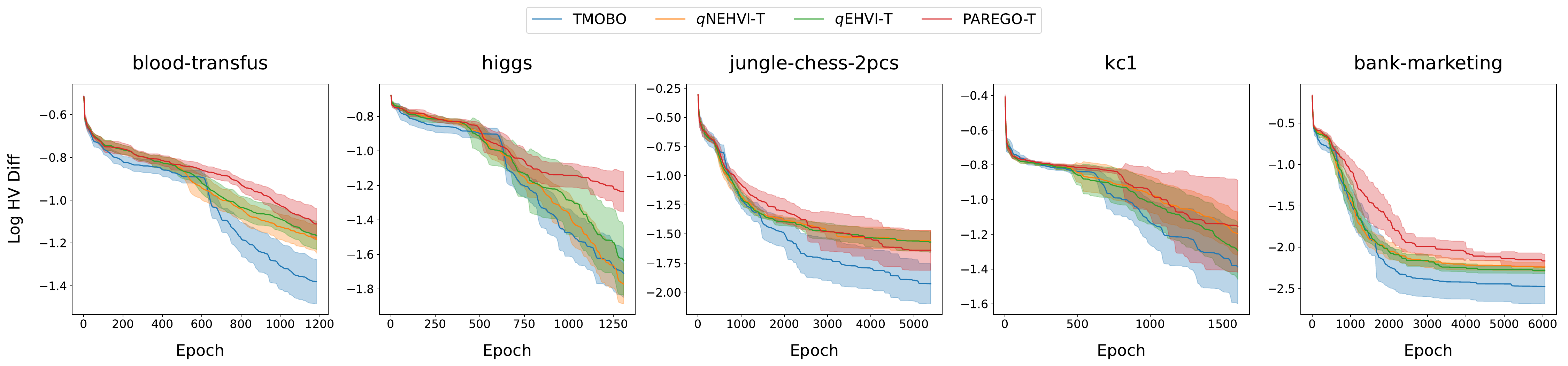}
    \caption{Average log Hypervolume difference against the number of epochs for each algorithm on five different hyperparameter tuning tasks (with first objective being validation accuracy and the second objective validation cross entropy). Each algorithm runs for 10 independent trials. The shaded region indicates two standard errors of the mean.}
    \label{fig-ByEpochSurrogate(B)}
\end{figure}

Figure \ref{fig-ByEpochSurrogate(B)} shows the performance of TMOBO and three alternative algorithms in terms of the average log HV difference against the total number of epochs. As the trajectory characteristics become more complex, each algorithm takes more effort to converge. However, TMOBO consistently outperforms the alternatives in both convergence speed and final HV difference, except for the second task on ``higgs'' where TMOBO and qNEHVI-T demonstrate comparable performance. Despite the changes in objective complexity, TMOBO maintains its advantage over the alternatives by leveraging information from partially trained models and dynamically adjusting training durations. This capability allows it to efficiently navigate the more challenging optimization tasks.

\subsection{Impact of Noisy Levels}\label{appx-result-noise}

\begin{figure}[H]
    \centering
    \includegraphics[width=\textwidth]{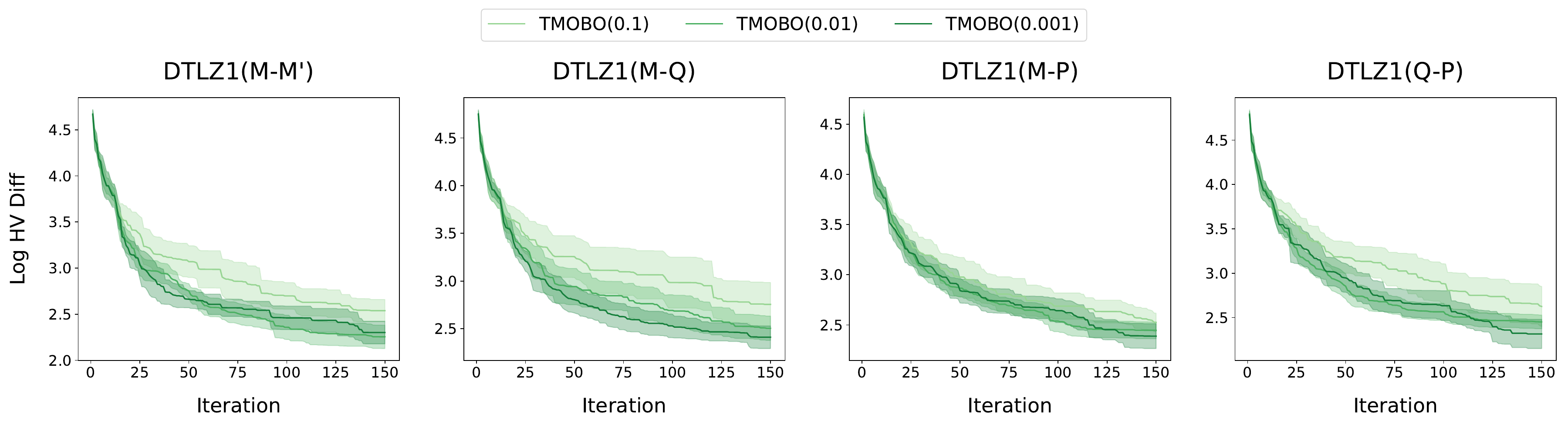}
    \caption{Average log Hypervolume difference against iterations for TMOBO on synthetic problems generated from DTLZ1 with noise level 0.1, 0.01, and 0.001. For each noise level, TMOBO runs for 10 independent trials. The shaded region indicates two standard errors of the mean.}
    \label{fig-ByIterSimulator(noise)}
\end{figure}

We then investigate the impact of stochasticity on the performance of TMOBO. To this end, for each synthetic problem generated from DTLZ, we add to each objective a Gaussian noise with standard deviations of 10\%, 1\%, and 0.1\% of the objective range. Figure \ref{fig-ByIterSimulator(noise)} highlights TMOBO’s adaptability and performance across different noise scenarios.

At lower noise levels (0.1\% and 1\%), TMOBO demonstrates stable and consistent performance. The algorithm quickly converges to a low HV difference, showcasing its ability to reliably predict trajectories and identify Pareto-optimal trade-offs. As the noise level increased to 10\%, the optimization becomes more challenging due to the amplified variability in objective evaluations, which TMOBO inherently inherits as part of the trade-off modeling. Despite these challenges, TMOBO continues to converge and achieves reasonably low HV differences by the end of trials.  

\subsection{Impact of Replication Strategy}\label{appx-resampling}

As a result of the analysis on noise levels, replicated observations at a query $(\bm{x}, t)$ might be needed to achieve adequate accuracy, especially in the presence of significant noises. While this particular scenario is not the focus of our current study, we outline a straightforward method to extend the proposed TMOBO to accommodate the requirement for replicated observations when the computational budget is sufficient. It is essential to recognize that with each noisy observation obtained at $(\bm{x}, t)$, the training process must start from scratch and will also output noisy observations from $(\bm{x}, 1)$ up to $(\bm{x}, t - 1)$. As a consequence, careful maintenance of multiple training procedures for replications is needed to ensure algorithmic efficiency. The pseudo-code for TMOBO-$P$, a modified version of TMOBO, is presented in Algorithm \ref{algo-TMOBO-P}. 

With an additional input $P$ denoting the number of replications, TMOBO-$P$ intends to include $P$ replicated observations at any visited query pair to mitigate the influence of noises. In each iteration, different from TMOBO, TMOBO-$P$ concurrently tracks $P$ independent training procedures for the ML model with the same sampled hyperparameter setting $\bm{x}'$ and ensures their progress remains consistent at the same epoch. Therefore, upon visiting a query pair $(\bm{x}, t')$, we can obtain $P$ replicated observations of it. Intuitively, the collection of sample means $\{\sum_{p = 1}^P\bm{y}_p(\bm{x}', 1) / P, \dots, \sum_{p = 1}^P\bm{y}_p(\bm{x}', t_{max}) / P\}$ form a compressed trajectory of $\bm{x}'$, i.e., average of $P$ observed trajectory of $\bm{x}'$. Then, the early stopping mechanism is applied to the compressed trajectory to determine when to stop all $P$ training procedures at the same time. Since the replicated training procedures are independent, TMOBO-$P$ can leverage parallel computing by assigning each training procedure to a specific processor. 

\begin{algorithm}[H]
    \caption{Framework of TMOBO-$P$}\label{algo-TMOBO-P}
    \textbf{Input:} Initial sets of inputs $Z$ and observations $Y$, number of replications $P$, and initial Pareto-optimal front $F$ identified from $Y$.
    \begin{algorithmic}[1]
        \While{the computational budget has not been exceeded}
            \State Fit $k$ GP models with $\bm{\mu}$ and $\bm{\Sigma}$ based on sets $Z$ and $Y$. 
            \State Sample a new $\bm{x}'$ by maximizing the TEHVI acquisition function. 
            \State Initialize $Z' \leftarrow \emptyset$ and $Y' \leftarrow \emptyset$.
            \For{$t' = 1$ \textbf{to} $t_{\max}$}
                \For{$p = 1$ \textbf{to} $P$}
                    \State Continue model training for the $t'$-th epoch to obtain observation $\bm{y}_p(\bm{x}', t')$ on the $p$-th processor.
                \EndFor
                \State Let $Z' \leftarrow Z' \cup \{(\bm{x}', t')\}$ and $Y' \leftarrow Y' \cup \{\sum_{p = 1}^P\bm{y}_p(\bm{x}', t') / P\}$ and update front $F$.
                \State Fit $k$ GP models with $\bm{\mu}$ and $\bm{\Sigma}$ based on sets $Z \cup Z'$ and $Y \cup Y'$.
                \If{EarlyStopping$(\bm{x}', t', \bm{\mu}, \bm{\Sigma}, F)$ is triggered} 
                    \State Break;
                \EndIf
            \EndFor
            \State Augment $Z'$ and $Y'$ into $Z$ and $Y$ respectively.
        \EndWhile
    \end{algorithmic}
\end{algorithm}

\begin{figure}[H]
    \centering
    \includegraphics[width=\textwidth]{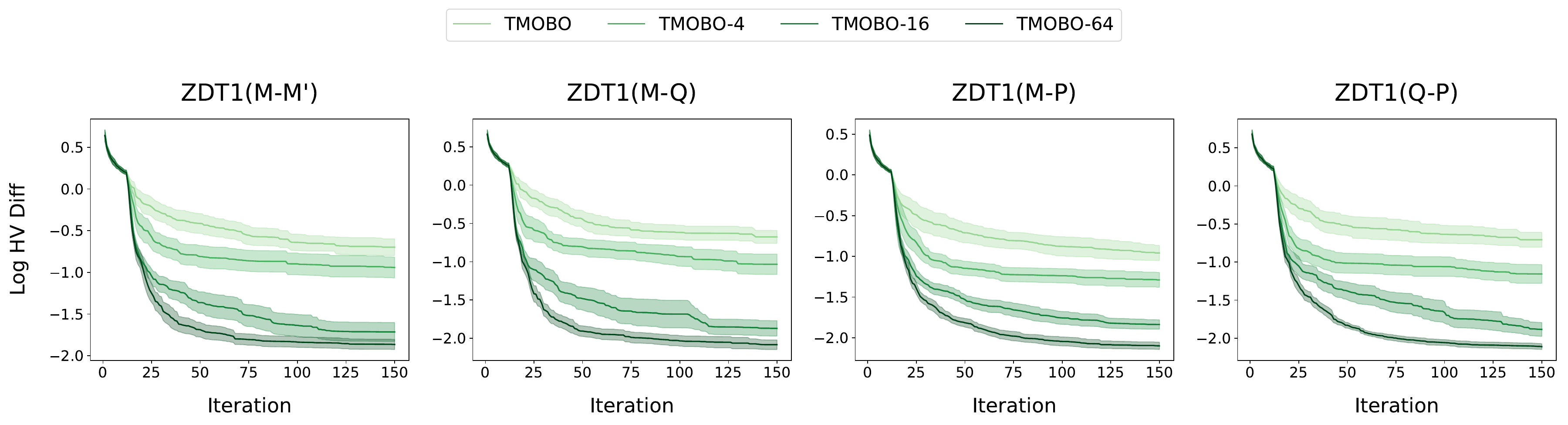}
    \caption{Average log Hypervolume difference against iterations for each algorithm on synthetic problems generated from ZDT1 with $10\%$ noise level. Each algorithm runs for 20 independent trials. The shaded region indicates two standard errors of the mean.}
    \label{fig-ByIterSimulatorReplication}
\end{figure}

Figure \ref{fig-ByIterSimulatorReplication} shows the performance of TMOBO-$P$ with $P = 1$, $4$, $16$, and $64$ on the synthetic problems derived from ZDT1. TMOBO proposed in the main paper can be considered a special case of TMOBO-$P$ with $P = 1$. This time, we add Gaussian noise to each objective with a standard deviation of $10\%$ of the objective range in order to emphasize the influence of noises. It can be observed that the performance of TMOBO-$P$ improves significantly in terms of the HV difference as more replications are allowed. This improvement is evident not only in rapid convergence during the initial stages but also in obtaining high-quality results at the end when comparing TMOBO with $P \geq 16$ to TMOBO with $P < 16$. In the meantime, benefiting from a large number of replications, TMOBO-$64$ has stable performance and its standard error is relatively small.

\end{document}